\documentclass[final,3p,times, sort&compress]{elsarticle}

\usepackage{amssymb}
\usepackage[printonlyused,nolist]{acronym}

\usepackage{amsmath}
\usepackage{tikz}

\usepackage{booktabs}
\usepackage{multirow}
\usepackage{rotating}

\usepackage{dsfont}

\newcommand{\N}{\mathds N}\newcommand{\R}{\mathds R}

\usepackage{array}
\newcolumntype{P}[1]{>{\centering\arraybackslash}p{#1}}
\newcolumntype{M}[1]{>{\centering\arraybackslash}m{#1}}

\usepackage[                                    
bookmarksopen=true,	
bookmarksnumbered=true,                 
plainpages=false,
pdfpagelabels,
colorlinks=true,
]{hyperref} 

\newcommand{\change}[1]{#1}

\DeclareMathOperator*{\argmax}{arg\,max}

\newcommand{\ie}{i.\,e.\;}
\newcommand{\eg}{e.g.\;}

\newcommand{\sub}[2][]{_{\mathrm{#2}#1}}

\newcommand{\Matrix}[1]{\mathbb{\MakeUppercase{#1}}}
\newcommand{\Set}[1]{\mathcal{\MakeUppercase{#1}}}
\newcommand{\VSpace}[1]{\mathrm{\MakeUppercase{#1}}}
\newcommand{\Rv}[1]{\MakeUppercase{#1}}

\newcommand{\nfeat}{n} % features / dimensions
\newcommand{\nsamp}{m} % samples
\newcommand{\iter}{t}

\newcommand{\ndim}{\nfeat}

\newcommand{\mmax}[1][]{\nsamp\sub[#1]{max}}
\newcommand{\minit}[1][]{\nsamp\sub[#1]{0}}
\newcommand{\mcand}[1][]{\nsamp\sub[#1]{cand}}
\newcommand{\mtest}[1][]{\nsamp\sub[#1]{test}}

\newcommand{\dmin}{d\sub{min}}  % closest distance
\newcommand{\eloo}[1][]{e\sub[#1]{LOO}}
\newcommand{\elooh}[1][]{\hat{e}\sub[#1]{LOO}}

\newcommand{\mmse}{\sub{MM}}
\newcommand{\wmmse}{\sub{WM}}
\newcommand{\mepe}{\sub{MEP}}
\newcommand{\eigf}{\sub{EI}}
\newcommand{\meigf}{\sub{GS}}
\newcommand{\ggess}{\sub{GG}}
\newcommand{\tead}{\sub{TE}}
\newcommand{\masa}{\sub{MA}}
\newcommand{\dlased}{\sub{DL}}

\newcommand{\smallcap}[1]{\textsc{#1}}
\newcommand{\acrofont}[1]{\textsc{#1}}

\newcommand{\Bhsingle}{\smallcap{HumpSingle}}
\newcommand{\Bhtwo}{\smallcap{HumpTwo}}
\newcommand{\Bgram}{\smallcap{GramLee}}
\newcommand{\Bbeker}{\smallcap{BekerLogan}}
\newcommand{\Begg}{\smallcap{Eggholder}}
\newcommand{\Bhimm}{\smallcap{Himmelblau}}
\newcommand{\Bbran}{\smallcap{Branin}}
\newcommand{\Bdrop}{\smallcap{DropWave}}
\newcommand{\Bishi}{\smallcap{Ishigami}}
\newcommand{\Bhart}{\smallcap{Hartmann}}

\newcommand{\Brose}[2][D]{\smallcap{Rosenbrock#2#1}}
\newcommand{\Back}[2][D]{\smallcap{Ackley#2#1}}
\newcommand{\Bmicha}[2][D]{\smallcap{Michalewicz#2#1}}
\newcommand{\Bschw}[2][D]{\smallcap{Schwefel#2#1}}
\newcommand{\Bstyb}[2][D]{\smallcap{StyblinskiTang#2#1}}

\newcommand{\tabspace}{3pt}

\definecolor{bblue}{HTML}{4878D0}
\definecolor{lightbrown}{HTML}{D5BB67}
\definecolor{bgreen}{HTML}{6ACC64}
\definecolor{bred}{HTML}{D65F5F}

\newcommand{\Circle}[2][1]{\tikz[baseline=-0.55ex]\draw[#2, fill=#2, opacity=#1,radius=.7ex] (0,0) circle ;}
\newcommand{\Rect}[2][1]{\tikz[]\draw[#2, fill=#2, opacity=#1] (0,0) rectangle ++(0.2,0.2) ;}
\newcommand{\SRect}[2][1]{\tikz[]\draw[#2, fill=#2, opacity=#1] (0,0) rectangle ++(0.16,0.16) ;}
\newcommand{\SDiamond}[2][1]{\tikz[baseline=0.1ex]\draw[#2, fill=#2, opacity=#1, rotate=45, rounded corners=0.1ex] (0,0) rectangle ++(0.14,0.14) ;}

\journal{Computer Methods in Applied Mechanics and Engineering}

\begin{document}

%% define acronyms
\begin{acronym}
	\acro{ml}[ML]{machine learning}
	\acro{ann}[ANN]{\acrofont{Artificial Neural Network}}
	\acro{pnn}[PNN]{\acrofont{Probabilistic Neural Network}}
	\acro{dnn}[DNN]{\acrofont{Deep Neural Network}}
	\acro{pinn}[PINN]{\acrofont{Physics-Informed Neural Network}}
	\acro{doe}[DoE]{design of experiments}
	\acro{lhs}[LHS]{Latin hypercube sampling}
	\acro{dgcn}[DGCN]{\acrofont{Deep Gaussian Covariance Network}}
	\acro{gp}[GP]{\acrofont{Gaussian Process Regression}}
	\acro{svgp}[SVGP]{\acrofont{Sparse Variational Gaussian Process Regression}}
	\acro{mcmc}[MCMC]{Markov-Chain Monte Carlo}
	\acro{vi}[VI]{variational inference}
	\acro{loocv}[LOOCV]{leave-one-out cross validation}
	\acro{bma}[BMA]{\textit{Bayesian model average}}
	\acro{iqd}[IQD]{interquartile distance}
	\acro{eigf}[EIGF]{\acrofont{Expected Improvement for Global Fit}}
	\acro{guess}[GUESS]{\acrofont{Gradient and Uncertainty Enhanced Sequential Sampling}}
	\acro{mepe}[MEPE]{\acrofont{Maximizing Expected Prediction Error}}
	\acro{mmse}[MMSE]{\acrofont{Maximum Mean Square Error}}
	\acro{wmmse}[wMMSE]{\acrofont{Weighted Maximum Mean Square Error}}
	\acro{tead}[TEAD]{\acrofont{Taylor-Expansion based Adaptive Design}}
	\acro{masa}[MASA]{\acrofont{Mixed Adaptive Sampling Algorithm}}
	\acro{ggess}[GGESS]{\acrofont{Gradient and Geometry Enhanced
		Sequential Sampling}}
	\acro{dlased}[DL-ASED]{\acrofont{Discrepancy Criterion and Leave One Out Error-based Adaptive Sequential Experiment Design}}
	\acro{pde}[PDE]{partial differential equation}
	\acro{bo}[BO]{Bayesian optimization}
	\acro{pca}[PCA]{principal component analysis}
	\acro{pdf}[PDF]{probability density function}
\end{acronym}

\begin{frontmatter}

\title{Gradient and Uncertainty Enhanced Sequential Sampling for Global Fit}

 \affiliation[ZF]{
 	organization={ZF Friedrichshafen AG},
	addressline={Graf-von-Soden-Platz 1},
	city={Friedrichshafen},
	postcode={88046},
	state={Baden-Wuerttemberg},
	country={Germany}}

\affiliation[HSNR]{
	organization={Institute of Modelling and High-Performance Computing, Niederrhein University of Applied Sciences},
	addressline={Reinarzstr. 49},
	city={Krefeld},
	postcode={47805},
	state={North Rhine-Westphalia},
	country={Germany}}

\affiliation[ZA]{
	organization={Zalando SE},
	addressline={Valeska-Gert-Straße 5},
	city={Berlin},
	postcode={10243},
	country={Germany}}

\affiliation[PI]{
	organization={PI Probaligence GmbH},
	addressline={Technology Centre Augsburg, Am Technologiezentrum 5},
	city={Augsburg},
	postcode={86159},
	state={Bavaria},
	country={Germany}}

\cortext[cor1]{Corresponding author}
\cortext[cor2]{Work done while at Institute of Modelling and High-Performance Computing}

\author[ZF,HSNR]{Sven Lämmle\corref{cor1}}
\ead{sven.laemmle@zf.com}
\author[ZA]{Can Bogoclu\corref{cor2}}
\ead{can.bogoclu@zalando.de}
\author[PI]{Kevin Cremanns}
\ead{kevin.cremanns@hs-niederrhein.de}
\author[HSNR]{Dirk Roos}
\ead{dirk.roos@hs-niederrhein.de}

\begin{abstract}
Surrogate models based on machine learning methods have become an important part of modern engineering to replace costly computer simulations. The data used for creating a surrogate model are essential for the model accu\-racy and often restricted due to cost and time constraints. Adaptive sampling strategies have been shown to reduce the number of samples needed to create an accurate model. This paper proposes a new sampling strategy for global fit called \smallcap{Gradient and Uncertainty Enhanced Sequential Sampling} (GUESS). The acquisition function uses two terms: the predictive posterior uncertainty of the surrogate model for exploration of unseen regions and a weighted approximation of the second and higher-order Taylor expansion values for exploitation. Although various sampling strategies have been pro\-posed so far, the selection of a suitable method is not trivial. Therefore, we compared our proposed strategy to 9 adaptive sampling strategies for global surrogate modeling, based on 26 different 1 to 8-dimensional deterministic benchmarks functions. Results show that GUESS achieved on average the highest sample efficiency com\-pared to other sur\-ro\-gate-based strategies on the tested examples. An ablation study considering the \change{behavior of GUESS in higher dimensions and} the importance \change{of} sur\-ro\-gate choice is also presented. 
\end{abstract}


\end{frontmatter}

\section{Introduction}\label{sec1}
Computer simulations have become an essential part of engineering and are used for various applications (\eg de\-sign optimization, structural mechanics, material science, etc.). Most of these applications require computationally demanding simulation models. \Ac{ml} models have emerged to mitigate the computational cost and are used as surrogates. For this task, \ac{ml} models learn the relationship between the response of the expensive simu\-lations from a dataset containing past observations. Various \ac{ml} methods were proposed as surrogate models (sometimes referred as response surface methods)\change{, \eg as solvers for \acp{pde} describing the behavior of high-dimensional vector spaces or fields \citep{karniadakis2021, lu2021b, li2021}, or probabilistic approximators of parametric response functions \citep{forrester2008}}. Among these types of surrogate models, \ac{gp} \citep{rasmussen2006} is a popular choice \citep{kopsiaftis2019, sterling2015} because of its flexibility and ability to predict the model uncertainty. However, drawbacks are the selection of a suitable covariance (kernel) function that is application dependent, and the computational burden for larger data sets. 
Various extensions to \acp{gp} such as the deep \acp{gp}\citep{damianou2013}, sparse \acp{gp} based on variational inference \citep{titsias2010, hensman2013}, and efficient matrix decomposition \citep{wang2019} were developed to overcome some of these limitations. One particularly promising approach is the combination of \acp{gp} with \acp{ann} to \acp{dgcn} \citep{cremanns2017, cremanns2021} to learn the non-stationary hyperparameters of the \ac{gp} together with combinations of different covariance functions. 

\acp{ann} alone can handle large datasets with high-dimensional inputs but lack the ability to predict the model uncertainty, unlike \acp{gp}. However, different approaches were developed in the past to make \acp{ann} uncertainty-aware based on dropout \citep{gal2016}, Bayesian inference \citep{mackay1992, lampinen2001, titterington2004, neal1996}, or ensembles \citep{lakshminarayanan2017, chua2018}.

The \ac{doe} \citep{sacks1989}, \ie the choice of samples used for training the surrogate model, is essential to create a globally accurate representation. In industrial applications, the amount of data is limited due to time and cost constraints. Thus, sampling points should be selected to improve the surrogate model accuracy as much as possible. Typically, the best choice of samples is unknown beforehand, and different methods are developed to overcome this problem. One widely used approach is \ac{lhs} \citep{mckay1979} where a \ac{doe} is created in advance based on sampling from the partitioned input space. Different modifications were proposed to reduce the correlation between the simulated variables and to ensure a space-filling design, \eg based around singular-value decomposition \citep{roos2016} or simulated annealing \citep{huntington1998}. This approach can be referred to as one-shot sampling since the \ac{doe} is created at once.
In contrast, \textit{adaptive} sampling techniques try to select new points sequentially. This allows using knowledge from previous iterations represented by the existing surrogate model to guide the adaptive sampling process. In the context of optimization using probabilistic models, this is also referred to as \ac{bo} \citep{mockus1994, jones1998}. However, this work focuses on the creation of \textit{globally accurate} surrogate models within the design domain, \ie the space of interest in which the surrogate model will be eventually used, based on adaptive sampling strategies. For this objective, an auxiliary function called acquisition is optimized to identify new experiments one by one. A popular choice for the acquisition function is the variance of the predictive distribution of the surrogate model, which is known within the context of Kriging \citep{krige1951} as \ac{mmse} \citep{sacks1989}. New points would then be obtained by selecting the sample with the highest predicted uncertainty. However, an effective adaptive sampling approach should consider two goals simultaneously, as pointed out in \citep{crombecq2011, liu2015}:
\begin{enumerate}
	\item[] \textit{Global Exploration} aims to discover the whole design domain and selects samples \eg with maximum distance to each other.
	\item[] \textit{Local Exploitation} selects samples at strategic points, \eg in regions with large prediction errors, that are important to capture the full function behavior.
\end{enumerate}

Various methods have been proposed, that extend the idea of using the predictive variance for global exploration and encourage local exploitation based on: \ac{loocv} error of the model  \citep{liu2017, kyprioti2020}, squared response difference \citep{lam2008} or gradient information \cite{chen2021}. \citep{fang2019} combined an adaptive weighted exploration based on triangulation with \ac{loocv} based exploration. Other approaches like the \ac{tead} \citep{mo2017} combine gradient information with distance based exploration. Another approach is to use the variance among a committee of models to formulate the exploitation criterion \citep{eason2014}. \change{See \cite{liu2018, fuhg2021} for a comprehensive list.}

\change{Related to the investigated strategies are adaptive sampling methods proposed for solving \acp{pde} in the context of physics-informed machine learning \citep{karniadakis2021}, especially for \acp{pinn} \citep{wu2023, tang2023, gu2021}. These methods improve the adaptive sampling procedure by minimizing the variance of the residuals between the surrogate approximation and the numerical solution. Those adaptive methods need additional evaluations of the \ac{pde} to refine the dataset used for training the surrogate model and require access to some information about the \acp{pde} describing the problem, in contrast to the strategies considered in this study. An overview is given in \citep{wu2023}.}

\change{Sample efficiency of the different adaptive strategies may vary depending on the response surface, \ie some methods may perform better at approximating specific response surfaces and worse at others. Selecting the best performing sampling method for the task at hand from the multitude of available algorithms is therefore challenging, since their performance is unpredictable beforehand. Hence}, a large scale comparative study is necessary to investigate their differences. So far, com\-para\-tive studies only consider a subset of the algorithms used in this work and mostly for fewer benchmarks  \citep{kupresanin2011, fuhg2021, mackman2013}; but for most adaptive sampling methods, the only empirical evaluation is offered by the original work introducing the method (see \eg \citep{chen2021, fang2019, kyprioti2020}). \change{Therefore}, this paper compares some of the recently developed adaptive sampling strategies for global surrogate modeling, based on 1 to 8-dimensional benchmark functions\footnote[1]{\label{github}source code: \url{https://github.com/SvenL13/GALE}}. In addition to the study, a novel ac\-qui\-sition function inspired by \ac{eigf} \citep{lam2008} and \ac{tead} \citep{mo2017} is presented, which is called \acf{guess}. In our study we show, that \ac{guess} provides promising results and an improvement regarding the sample efficiency over the inspired strategies based on our experiments. 

The paper is structured as follows: Section \ref{sec2} starts with the theoretical background, introducing the deployed ML algorithms for surrogate modeling. Section \ref{sec3} introduces the investigated adaptive sampling strategies. In Section \ref{sec4}, the results of the benchmark study are presented. Finally, in Section \ref{sec5}, concluding remarks are given.
\change{Appendices contain complementary results for higher-dimensional problems (\ref{secA_1}), as well as an ablation study regarding the model choice (\ref{secA_models}), computational effort (\ref{secA_comp}), implementation details (\ref{secB}), and the used benchmark functions (\ref{secC}).}

\section{Theoretical Background}\label{sec2}

\subsection{Surrogate Modeling}\label{sec2_1}
For a response function $ f\colon \R^n \rightarrow \R; \textbf{x} \mapsto f(\textbf{x})=y$ of a simulation model  with input $\textbf{x}$ and output $y$, the surrogate model $\hat{f}$ approximates the relationship $\hat{f}(\textbf{x}^*; \Set{D})\approx y^*$ based on the training data $\Set{D}=\{ (\textbf{x}_i, y_i) \vert 1 \leq i \leq m, i \in \N \}$, where $\textbf{x}_i$ is the $i$-th row in $\textbf{X}_{i, :}$, $\textbf{X} \in \R^{m \times n}$, $\textbf{y} \in \R^m$ with $m$ and $n$ denoting the number of samples and the number of input parameters, respectively. The indices $(i, : )$  represent the entry in the $i$-th row of the sample matrix. Testing quantities $\left(\textbf{x}^*\notin \Set{D}\right)$ are indicated with the asterisk (*) superscript.

\change{Four} probabilistic surrogate models are considered in this work: \ac{gp}, \change{\ac{svgp} \citep{hensman2013}}, \ac{dgcn} and an ensemble of \acp{pnn} \citep{neal1996, lakshminarayanan2017, chua2018}. The prediction of a probabilistic model is a random variable $\hat{f}(\textbf{x}; \Set{D}) = \Rv{Y} \sim p(\tilde{y} \vert \textbf{x}, \Set{D})$ and the value $\tilde{y}$ used for the further analysis is often the mean $\mu_{\Rv{Y}}$ or a sample from $\Rv{Y}$. Later, the variance of $Y$ can be utilized to guide the adaptive sampling process to regions where the model has high uncertainty about the possible outcome. In the following a short description of \acp{gp} is given. \change{\ac{svgp} is introduced in \ref{secA_1_3}, while} \ac{dgcn} and \ac{pnn}, together with the results, can be found in \ref{secA_models}.

\subsection{Gaussian Process}\label{sec2_1_2}
A \ac{gp} is a distribution over functions, which can be represented as a collection of infinite number of random var\-iables, a finite number of which follow a joint Gaussian distribution. The choice of kernel function influences the prediction capability of the \ac{gp}. A common correlation function is the Mat\'ern kernel \citep{stein1999, matern1986} which determines the covariance for a pair of points $\textbf{x}$ and $\textbf{x}'$:
\begin{equation}
	k(\textbf{x}, \textbf{x}', l) = \dfrac{2^{1- \nu}}{\Gamma (\nu)} \left( \dfrac{\sqrt{2\nu} r}{l} \right) ^{\nu} K_{\nu} \left( \dfrac{\sqrt{2\nu} r}{l} \right)\sigma^2_{f} \label{eq7}
\end{equation}
where $r=\Vert \textbf{x}-\textbf{x}' \Vert_2$ is the Euclidean distance, $K_{\nu}(\cdot)$ is a modified Bessel function of the second kind \citep{abramowitz1972}, $\Gamma(\cdot)$ is the gamma function, $\sigma\sub{f} ^2$ is the kernel variance and $l$ is the learnable length scale. $\nu$ is a positive parameter with popular values $3/2$ or $5/2$. 
To train the model parameters $\boldsymbol{\theta}\sub{GP}=\{l, \sigma\sub{n}^2, \sigma\sub{f}^2\}$ we can maximize the marginal log-likelihood
\begin{align}
	\log p(\textbf{y} \vert \textbf{X}, \boldsymbol{\theta}\sub{GP}) = - \dfrac{m}{2} \log(2\pi) 
	-\dfrac{1}{2} \log \vert \textbf{K}\sub{N} \vert 
	- \dfrac{1}{2}\textbf{y}^T \textbf{K}\sub{N} ^{-1} \textbf{y}
	 \label{eq8}
\end{align}
where $\sigma\sub{n}^2$ is the noise variance and $\textbf{K}\sub{N} =\textbf{K} + \sigma\sub{n} ^2 \textbf{I}$ is the covariance matrix, with $\textbf{K}_{i,:}=\left[k(\textbf{x}_i, \textbf{x}_1, l),  ..., k(\textbf{x}_i, \textbf{x}_m, l)\right]$ and $\textbf{I}$ as the identity matrix. In contrast to the maximum-likelihood estimate, it is also possible to use a Bayesian approach for parameter estimation (see \eg\citep{hensman2015}).
Since the \ac{gp} is defined as a joint Gaussian distribution, the prediction at a point $\textbf{x}^*$ is a Gaussian variable $\Rv{Y}^*$ with mean $\mu _{Y^*}$ and variance $\sigma_{Y^*}^{2}$:
\begin{align}
	\hat{f}\sub{GP}( \textbf{x}^* ; \boldsymbol{\theta}\sub{GP}) = \mu _{Y^*} = \textbf{k}^{T} \textbf{K}\sub{N}^{-1} \textbf{y} \label{eq9}
\end{align}
\begin{align}
	\mathbb{V}\left[\hat{f}\sub{GP}(\textbf{x}^*; \boldsymbol{\theta}\sub{GP})\right] = \sigma_{Y^*}^{2} =  
	k( \textbf{x}^*, \textbf{x}^*, l) - \textbf{k}^T \textbf{K}\sub{N}^{-1} \textbf{k} \label{eq10}
\end{align}
where $\textbf{k}$ denotes the vector of correlations between $\textbf{x}^*$ and the design points, $\textbf{k} = \left[k(\textbf{x}^*, \textbf{x}_1, l),  ..., k(\textbf{x}^*, \textbf{x}_m, l)\right]^T$.

\subsection{Leave-One-Out Cross Validation}\label{sec2_2}
The $k$-fold cross validation \citep{kohavi1995} can be used to estimate the surrogate model accuracy on unseen samples. In the special case for $m$ folds, we obtain the \ac{loocv}, where $m$ observations are made. For every $i \in [1, m]$, a model is trained on the $m-1$ observations in order from the reduced set $\Set{D}_{\neg,i}=\Set{D} \backslash \{\mathbf{x}_i, y_i\}$. Then the \ac{loocv} error at a location $\mathbf{x}_i$ can be calculated with the squared error loss as
\begin{align}
	\eloo^2(\mathbf{x}_i; \hat{f}, \Set{D})=& \left( \tilde{y}_i - \tilde{y}_{\neg, i} \right)^2, \quad \forall i \in [1, m] \label{14}
\end{align}
where $\tilde{y}_i$ is the prediction at $\mathbf{x}_i$ from the model trained on the whole dataset $\Set{D}$ and $\tilde{y}_{\neg, i}$ is the prediction from the model trained on the reduced set $\Set{D}_{\neg,i}$.

\subsection{Fast Approximation for \ac{gp}}\label{sec2_2_1}
Given a set of training samples $\Set{D}$ and a \ac{gp} with mean function $\mu(x)=0$ and kernel function $k(\cdot)$, we can derive a closed-form solution for the approximated \ac{loocv} error  \citep{sundararajan2001, rasmussen2006}
\begin{equation}
	\elooh(\mathbf{x}_i)=\frac{\mathbf{\Lambda}_{i,:}\mathbf{y}}{\mathbf{\Lambda}_{ii}}\label{eq15}\nonumber
\end{equation}
where $\mathbf{\Lambda}$ is the precision matrix $\mathbf{\Lambda}=\mathbf{K}^{-1}$ and the subscript $ii$ denotes the
entry in the $i$-th row and $i$-th column of the matrix.
With the fast approximation of $\eloo$, the computational complexity can be reduced from $\mathcal{O}(m^4)$ to $\mathcal{O}(m^3)$, since the inverse of $\mathbf{K}$ is calculated only once, plus $\mathcal{O}(m^2)$ for the whole \ac{loocv} procedure \citep{rasmussen2006}.

\section{Adaptive Sampling Strategies for Global Fit}\label{sec3}
Considering an expensive black-box function $ f\colon \VSpace{X} \rightarrow \VSpace{Y}$ with output $y \in \VSpace{Y} \subseteq \R $, where the objective is to create a surrogate model $\hat{f}$ that is globally accurate within the design domain $\VSpace{X} \subseteq \R^n$. The aim of adaptive sampling strategies is to form a sequential procedure to suggest new sample points for improving the surrogate model accuracy as much as possible. 

A set of initial samples $\Set{D}_0=\{ (\textbf{x}^0_i, y^0_i) \vert 1 \leq i \leq m_0, i \in \N \}$ is generated first using a space filling sampling approach (\eg \ac{lhs}), where $\textbf{x}^0_i$ is the $i$-th row in $\textbf{X}^0_{i, :}$, $\textbf{X}^0 \in \VSpace{X}^{{m_0}}$, $\textbf{y}^0 \in \VSpace{Y}^{m_0}$ with $m_0$ denoting the number of initial samples.
The methods covered in this work employ an acquisition function $\phi\colon \VSpace{X} \rightarrow \R$ that is maximized in order to obtain a new sampling point in the $\iter$-th iteration
\begin{equation}
	\textbf{x}^{\iter}=\argmax_{\textbf{x}\in\mathrm{X}}\phi \left( \textbf{x}; \Set{D}_{\iter}, \hat{f}_{\iter} \right) \label{eq17}
\end{equation}
where $\hat{f}_{\iter}$ is the trained surrogate model using the dataset $\Set{D}_{\iter}$.
From here on, we drop writing the explicit dependence of $\phi$ on $\Set{D}_{\iter}$ and $\hat{f}_{\iter}$.

Some methods rely on selecting the candidate sample with the highest acquisition value over some candidate points $\textbf{x}^c\in\textbf{X}^c$ instead of finding the maximum using an optimization algorithm. The candidate set $\textbf{X}^c$ is then obtained by sampling (\eg random sampling, \ac{lhs}, ...) from $\VSpace{X}$. In this work, we have restricted our view to single-response problems. An overview of adaptive sampling for multi-response models is given in \cite{liu2018}.

\subsection{Adaptive Sampling Workflow}\label{sec3_1}
The adaptive sampling approach can be described in six steps:

\begin{enumerate}[Step 1:]
	\item Create the initial design points $\textbf{X}^0$. If needed\change{, create} the candidate set $\textbf{X}^c$ with a sampling method (\eg \ac{lhs}) and set $\iter=0$.
	\item Evaluate the expensive black-box function $f$ at $\textbf{X}^0$ to obtain the responses $\textbf{y}^0$ and construct the initial dataset $\Set{D}_0$.
	\item Train the surrogate model $\hat{f}_{\iter}$ using $\Set{D}_\iter$.
	\item Maximize the acquisition function $\phi$ with respect to the design domain $\VSpace{X}$ or the candidate set $\textbf{X}^c$ to identify a new design point $\textbf{x}^{\iter}$ (Eq.~\eqref{eq17}) using $\hat{f}_{\iter}$ and $\Set{D}_{\iter}$. 
	\item Evaluate the expensive black-box function $f$ at $\textbf{x}^{\iter}$ to receive the response $y^{\iter}$. Extend $\Set{D}_{\iter}$ with the new data to obtain the dataset $\Set{D}_{\iter+1}=\Set{D}_\iter \bigcup \left\{\left(\textbf{x}^{\iter}, y^{\iter} \right) \right\}$. \change{Create a new candidate set $\textbf{X}^c$ if needed}.
	\item Examine if a stopping condition is met (\eg maximum iterations, time constraint, accuracy goal). Otherwise increment $\iter$ and go to Step 3.
\end{enumerate}

\subsection{Acquisition Functions}\label{sec3_2}
Acquisition functions are constructed to guide the adaptive sampling process in regions that are most beneficial to generate new data. Previous research indicates that the most effective methods use a combination of exploration and exploitation strategies \citep{fuhg2021}. Additional adjustment factors may also be used to weigh between exploration and exploitation based on past observations. In the following, different methods are presented based on these criteria.

\textbf{MMSE:} The \acf{mmse} introduced for Kriging by Sacks and Schiller \citep{sacks1989} is a straightforward approach that is based on the predictive posterior variance of the surrogate model and is therefore purely exploration based
\begin{equation*}
	\phi\mmse (\textbf{x}) = \sigma^2_{Y}
\end{equation*}

\textbf{wMMSE:} The \acf{wmmse} \citep{kyprioti2020} is an extension to the \ac{mmse} that considers an additional exploitation criterion. The trade-off between exploration and exploitation is adjusted with an additional factor $\alpha\wmmse \in \R_+$, where $\alpha\wmmse>1$ favors exploration and  $\alpha\wmmse<1$ exploitation. On average, the authors provided empirical evidence using benchmark functions that $\alpha\wmmse=1$ yields the best performance. The acquisition function is defined as
\begin{equation*}
	\phi\wmmse(\textbf{x}) = \left(\gamma(\textbf{x})\right)^{\alpha\wmmse} \sigma^2_{Y}
\end{equation*}
where $\gamma(\cdot)$ contains information about the surrogate model error at the closest sample within the observed design points $\textbf{x}_o\in \textbf{X}^{\iter}\in\R^{m_t\times n}$
\begin{align*}
	\gamma(\textbf{x}):= \eloo^2(\textbf{x}_o), \; \textbf{x}\in \Set{V}_o
\end{align*}
where $\Set{V}_o$ is the Voronoi cell assigned to $\textbf{x}_o$
\begin{align*}
	\Set{V}_o = \left\{\textbf{x} \in \VSpace{X} \big\vert \Vert \textbf{x}-\textbf{x}_o \Vert_2 \leq \Vert \textbf{x} - \textbf{x}_j \Vert_2, \forall j \neq o \right\}
\end{align*}
where $o, j \in \{1,...,m_\iter\}$ and $m_\iter$ is the number of observed samples in the $\iter$-th iteration. Therefore, the Voronoi partitioning is used to make $\eloo^2$ available over the domain $\VSpace{X}$ according to the observed samples $\textbf{x}_o$.

\textbf{MEPE:} The \acf{mepe} \citep{liu2017} uses the cross-validation error $\eloo^2$ for local exploitation and the prediction variance $\sigma^2_{Y}$ for exploration. The acquisition function is defined as
\begin{align*}
	\phi\mepe(\textbf{x}) =& \alpha\mepe \gamma(\textbf{x}; \Set{D}_{\iter}) + (1-\alpha\mepe)\sigma^2_{Y}
\end{align*}
where the balance factor $\alpha\mepe \in \R^{[0, 1)}$ weighs exploration and exploitation adaptively, depending on the true and cross-validation error at the latest observed point
\begin{equation*}
	\alpha\mepe = \begin{cases}
		0.5, & \iter=0,\\
		0.99 \min \left( 0.5 
		\frac{\displaystyle e^2\sub{true} \left(\textbf{x}^{\iter-1}\right)}
		{\displaystyle \elooh^2 \left(\textbf{x}^{\iter-1}\right)}, 1 \right), & \iter>0 
	\end{cases}
\end{equation*}
where the true error $e^2\sub{true}$ is obtained from $f$ and $\hat{f}_{\iter-1}$ at the previous observed location $\textbf{x}^{\iter-1}$.

\textbf{EIGF:} Inspired by the Expected Improvement criterion for global optimization \citep{jones1998}, the \acf{eigf} \citep{lam2008} is adapted to combine information from the predicted variance together with the squared response difference as
\begin{equation}
	\phi\eigf(\textbf{x}) = \left( \hat{f}_{\iter}(\textbf{x}) - f(\textbf{x}_{o}) \right)^2   +\sigma^2_{Y}, \; \textbf{x}\in \Set{V}_o \label{eq24}
\end{equation}

\textbf{GGESS:} Chen et al. \citep{chen2021} modified the \ac{eigf} criterion and proposed the \acf{ggess}, by including additional gradient information to improve the acquisition function
\begin{align}
	\phi\ggess(\textbf{x}) = \left( f(\textbf{x}_{o}) - \hat{f}_{\iter}(\textbf{x}) - \nabla_{\textbf{x}}\hat{f}_{\iter}(\textbf{x})^T(\textbf{x}_{o} - \textbf{x}) \right)^2 +\sigma^2_{Y}, \; \textbf{x}\in \Set{V}_o \label{eq25}
\end{align}
where $\nabla_{\textbf{x}}\hat{f}_{\iter}(\textbf{x})$ is the approximate gradient vector of the true function $f$ at $\textbf{x}$.

\textbf{TEAD:} Similar to \ac{ggess}, \acf{tead} \citep{mo2017} uses the gradient vector $\nabla_{\textbf{x}}\hat{f}_{\iter}(\textbf{x})$ to find potential samples in regions of high interest. The exploitation criterion is constructed around a Taylor-based approximation of the second- and higher-order Taylor expansion values
\begin{equation}
	\delta(\textbf{x}) = \left\vert \hat{f}_{\iter}(\textbf{x}) - \hat{f}_{\iter}(\textbf{x}_o)-
	\nabla_{\textbf{x}}\hat{f}_{\iter}(\textbf{x}_o)^T(\textbf{x}-\textbf{x}_o) \right\vert, \; \textbf{x}\in \Set{V}_o\label{eq27}
\end{equation}
with $\nabla_{\textbf{x}}\hat{f}_{\iter}(\textbf{x}_o)^T(\textbf{x}-\textbf{x}_o)$ as the first-order Taylor expansion of the model at $\textbf{x}_o$.  $\nabla_{\textbf{x}}f$ is replaced with the faster approximation $\nabla_{\textbf{x}}\hat{f}$. The exploration is based on the closest distance between a point $\textbf{x}$ and the observed data points $\textbf{X}^\iter$
\begin{equation}
	\dmin(\textbf{x}) = \min_{\textbf{x}_i \in \textbf{X}^\iter} \left\Vert \textbf{x} - \textbf{x}_i \right\Vert_2 \label{eq28}
\end{equation}
The acquisition function for \ac{tead} is the combination of both criteria
\begin{equation*}
	\phi\tead(\textbf{x}) = \frac{\dmin(\textbf{x})}{\max\limits_{\textbf{x}^c \in \textbf{X}^c} \dmin(\textbf{x}^c)} + \alpha\tead(\textbf{x}) \frac{ \delta(\textbf{x})}{\max\limits_{\textbf{x}^c \in \textbf{X}^c} \delta(\textbf{x}^c)}
\end{equation*}
where $\alpha\tead \in \R^{[0, 1]}$ is an additional adjustment factor
\begin{equation*}
	\alpha\tead(\textbf{x}) = 1 - \dmin(\textbf{x}) / d\sub{max} 
\end{equation*}
with $d\sub{max}$ as the maximum distance of any two points in $\VSpace{X}$.

\textbf{MASA:} The \acf{masa} \citep{eason2014} uses a committee of different models $\Set{M} = \{ {\hat{f}^\Set{M}_1, ..., \hat{f}^\Set{M}_{n_\Set{M}}}\}$ to construct an exploitation criterion based on the variance among the $n_\Set{M}$ committee members
\begin{equation*}
	F\sub{QBC}(\textbf{x}; \Set{M}) = \frac{1}{n_\Set{M}} \sum_{i=1}^{n_\Set{M}} \left( \hat{f}_i^\Set{M}(\textbf{x}) - \overline{\hat{f}^\Set{M}}(\textbf{x}) \right)^2
\end{equation*}
where $\overline{\hat{f}^\Set{M}}(\textbf{x})=\frac{1}{n_\Set{M}}\sum_{i=1}^{n_\Set{M}} \hat{f}_i^\Set{M}(\textbf{x})$ is the average of the committee prediction.
Similar to \ac{tead}, the exploration is based on the minimal distance between the observed and candidate points (Eq.~\eqref{eq28}). The acquisition function is given as
\begin{equation*}
	\phi\masa(\textbf{x}; \Set{M}) =  
	\frac{F\sub{QBC} (\textbf{x}; \Set{M})}
	{\max\limits_{\textbf{x}^c \in \textbf{X}^c} F\sub{QBC} (\textbf{x}^c; \Set{M})}
	+ \frac{\dmin(\textbf{x})}
	{\max\limits_{\textbf{x}^c \in \textbf{X}^c} \dmin(\textbf{x}^c)} 
\end{equation*}

\textbf{DL-ASED:} The \acf{dlased} \citep{fang2019} uses a weighted combination of distance-based exploration together with \ac{loocv}-based exploration. 
The notion of support points is introduced for the exploration criterion. Therefore, $n+1$ support points $\textbf{X}^s=\left[\textbf{x}^s_0, ...,\textbf{x}^s_{n+1}\right]^T$, with $\textbf{x}_i^s \in \Matrix{X}^{\iter}$, are assigned to each can\-di\-date $\textbf{x}^c$, based on a triangulation scheme (see \cite{fang2019}). 

The exploration criterion is calculated as
\begin{equation}
	g(\textbf{x}) = v_{s}^p \prod_{i=1}^{n+1} (\Vert \textbf{x} - \textbf{x}_{i}^{s} \Vert_2 \cdot \Vert \hat{f}_{\iter}(\textbf{x}) - f(\textbf{x}_{i}^{s}) \Vert_2) \label{eq33}
\end{equation}
where $\textbf{x}_i^{s}$ is a local support point around $\textbf{x}$, $v_{s}$ is the volume of the polyhedron constructed with the support points $\textbf{X}^{s}$ and $p$ is a coefficient that is not further described in \citep{fang2019}. In the following $p=1$ is used. 
For the local exploitation based on $\eloo$, an additional model $\hat{f}_{\eloo}\colon \VSpace{X} \rightarrow \R$ is used to define a map over the entire $\VSpace{X}$. The model approximates the relationship $\hat{f}_{\eloo, \iter}(\textbf{x};
 \Set{D}_{\iter})\approx\eloo[,\iter]$ using the data set $ \left\{ (\textbf{x}_i, \eloo (\textbf{x}_i)) \ \big\vert \ \textbf{x}_i \in \textbf{X}^{\iter} \ \text{for} \ i = 0, ..., m_\iter \right\}$. The acquisition func\-tion is then defined as
\begin{align}
	\hspace*{-0.2cm}\phi\dlased(\textbf{x}) =
	(1 - \alpha\dlased) \frac{g(\textbf{x})}{\max\limits_{\textbf{x}^c \in \textbf{X}^c} g(\textbf{x}^c)}
	+ \alpha\dlased \frac{\hat{f}_{\eloo, \iter}(\textbf{x}; \Set{D}_{\iter})}{\max\limits_{\textbf{x}^c \in \textbf{X}^c} \hat{f}_{\eloo, \iter}(\textbf{x}^c; \Set{D}_{\iter})} \label{eq_aq_dlased},
\end{align}
where $\alpha\dlased$ is the adaptive weight in the  $\iter$-th iteration
\begin{equation*}
	\alpha\dlased= 
	\begin{cases}
		0.5, & \iter=0,\\
		\min \left( 0.5 \frac{\displaystyle \zeta\sub{local}}{\displaystyle \zeta\sub{global}}, 1 \right), & \iter>0
	\end{cases}
\end{equation*}
with $\zeta\sub{global} = \sum_{i=1}^{m_{\iter}} \eloo[,\iter]^2 \left(\textbf{x}_i\right) / \sum_{j=1}^{m_{\iter-1}} \eloo[,\iter-1]^2 (\textbf{x}_j)$ as the global and $\zeta\sub{local} = \eloo[,\iter]^2 \left(\textbf{x}^{\iter}\right) / \eloo[,\iter-1]^2 \left(\textbf{x}^{\iter}\right)$ as the local improvement of $\eloo$ in the $\iter$-th iteration. $\eloo[,\iter-1]^2$ is the \ac{loocv} error based on the previous model $\hat{f}_{\iter-1}$ and dataset $\Set{D}_{\iter-1}$.

\textbf{GUESS:} We pro\-pose a novel criterion, the \acf{guess}. The acquisition uses the predicted standard deviation for exploration of the design domain and a Taylor expansion based approximation of the second- and higher-order reminders for exploitation. The exploitation term is weighted by the predicted standard deviation, i.e. it acts as a penalty to avoid choosing samples too close to already observed samples. The acquisition function is given by
\begin{align}
	\phi\meigf(\textbf{x}) = \left(\delta(\textbf{x}) + 1 \right)\sigma_{Y}, \; \textbf{x}\in \Set{V}_o \label{eq26}
\end{align}
where $\delta(\textbf{x})$ is the Taylor-based approximation of the second- and higher-order Taylor expansion values (Eq.~\eqref{eq27}).
\begin{figure*}[!tb]
	\centering
	\includegraphics[width=1.0\textwidth]{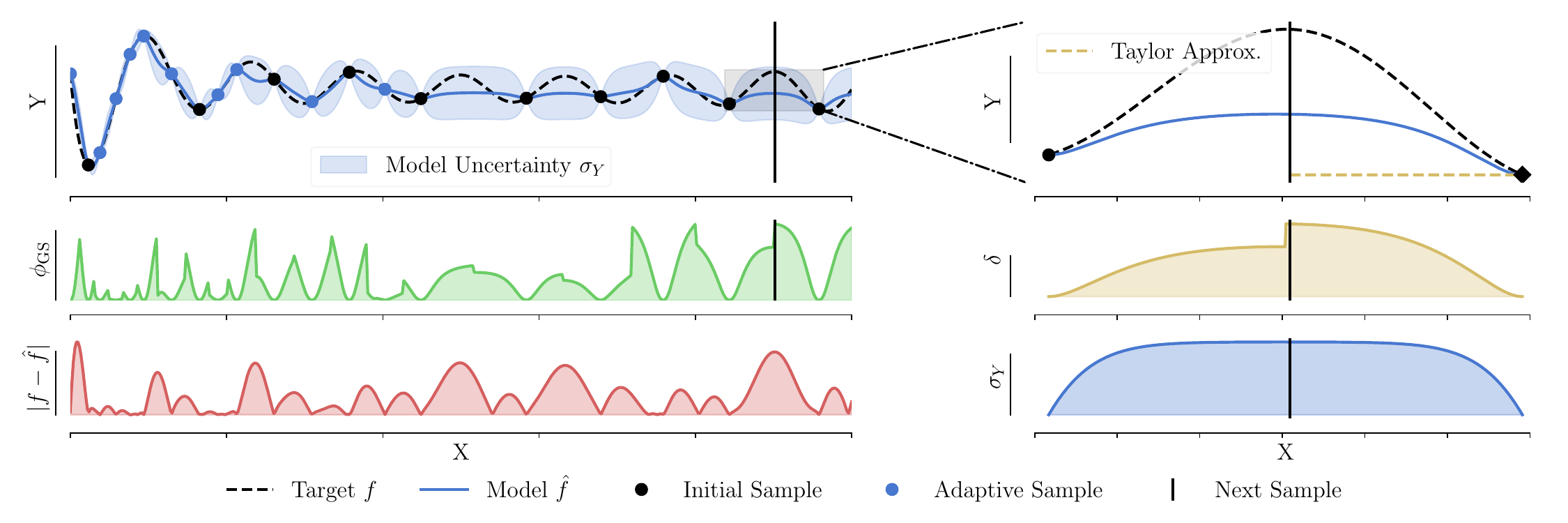}
	\caption{\change{Illustration of \ac{guess} with \ac{gp} surrogate model and Matérn kernel ($\nu=3/2$) for the $\Bgram$ function.
	The acquisition function $\phi\meigf$ (\SRect[0.3]{bgreen}) is decomposed into second and higher-order reminders $\delta$ (\SRect[0.3]{lightbrown}) and predicted standard deviation $\sigma_{Y}$ (\SRect[0.3]{bblue}). Plot in the top right corner shows the fist-order Taylor approximation at the closest observed point $\textbf{x}_o$ (\SDiamond{black}). 
	$\delta$ is large in areas with high non-linearity, defined as the difference between surrogate prediction $\hat{f}$ and first order approximation. The step in $\delta$ is due to the switch of expansion point, \ie another observed sample $\textbf{x}_o$ is closer. As reference, the absolute error between the underlying function and surrogate model is visualized (\SRect[0.3]{bred}).}}\label{fig_guess}
\end{figure*}
\change{Figure \ref{fig_guess} illustrates different components of \ac{guess}. We perform \ac{gp} using a Matérn kernel ($\nu=3/2$) on a toy dataset, where 10 points are sampled with \ac{lhs} (\Circle{black}) and 10 points are sampled with \ac{guess} (\Circle{bblue}) from the $\Bgram$ function. As can be seen in Figure \ref{fig_guess}, the second- and higher-order Taylor expansion values in Eq.~\eqref{eq27} act as a measure of non-linearity of the function (\Rect[0.3]{lightbrown}). 
The approximation accuracy of the first-order Taylor expansion decreases as the non-linearity of $f$ increases as well as the distance to the expansion point increases \citep{mo2017}.	
In most scenarios, $\phi\meigf$ is dominated by $\delta(\textbf{x})\sigma_{Y}$. However, in situations with small gradients, the addition of $\sigma_{Y}$ acts as a fallback to locate samples in regions with large model uncertainty.
Modifying Eq.~\eqref{eq26} by replacing $\delta(\textbf{x})$ with $\delta(\textbf{x})^{\alpha\meigf}$, where $\alpha\meigf\in\R_+$, controls the trade-off between exploration ($\alpha\meigf<1$) and exploitation ($\alpha\meigf>1$).
In the following, we considered \ac{guess} only with $\alpha\meigf=1$ which gives a natural trade-off, \ie model uncertainty of the model will decrease with increasing sample size.}

\subsection{Classification of Sampling Strategies}\label{sec3_3}
Table \ref{tab1} shows the classification based on the exploration and exploitation components for the different sampling strategies. Most of the presented acquisition functions are originally formulated for Kriging but a majority of them can be adapted without change to other probabilistic models like \acp{dgcn} or \change{\acp{pnn}}. However, some of the methods (\ac{mepe}, \ac{dlased} and \ac{wmmse}) rely on calculating the \ac{loocv} error, which can be a computationally demanding task for other surrogate models without a fast approximation available like described in Section \ref{sec2_2_1}.
\begin{table}[!ht]
		\centering
		\begin{minipage}{\columnwidth}
			\caption{Classification of the investigated sampling strategies.}\label{tab1}
			\vspace{\tabspace}
			\centering
			\begin{tabular}{@{}ccc@{}}
				\toprule
				\multirow{2}{*}{\textbf{Exploitation}} & \multicolumn{2}{c}{\textbf{Exploration}} \\
				& distance-based & predictive variance-based \\
				\midrule
				no exploitation				& \ac{lhs} & \ac{mmse} \\ \hline
				\multirow{2}{*}{\ac{loocv}-based}					&   
				\multirow{2}{*}{\ac{dlased}\footref{cand}  }
				& \ac{mepe} \\ 
				&   & \ac{wmmse}\footnote{\label{cand}Discontinuous acquisition function based on $\textbf{X}^c$} \\ \hline
				committee-based	  & \ac{masa}\footref{cand} &  - \\  \hline
				\multirow{3}{*}{geometry-based} &  & \ac{eigf} \\ 
				& \ac{tead}\footref{cand} & \ac{guess}\footref{cand} \\
				& & \ac{ggess}\footref{cand} \\
				\bottomrule
			\end{tabular}
		\end{minipage}
\end{table}

\section{Benchmark Study}\label{sec4}
A benchmark study using analytical functions was conducted to compare the sampling methods described before. The results are presented in the following. Besides the adaptive strategies, one-shot sampling using \ac{lhs} was also investigated as a baseline.
\subsection{Test Scheme}\label{sec4_1}
The evaluation was based on 15 different deterministic benchmark functions chosen from optimization literature (see \ref{secC}). The benchmark functions
range from 1 to 8 dimensions and input dimensions were scaled to have unit length. The functions were selected to cover a wide palette of various surface structures (multiple local optima, valley shaped, bowl shaped, etc.) which propose different challenges that could occur in engineering problems. 

Initial samples, optimization and other aspects introduce stochasticity to the sampling process. Moreover, assessing the performance of an adaptive sampling strategy on a single run may be misleading due to randomness. Therefore, every benchmark function was repeated 10 times for each method and fixed seeds were used for each run to ensure reproducibility and similar numerical conditions between the evaluated methods. 

The acquisition functions in Section \ref{sec3_2} were maximized in each iteration to propose new points. Some of the investigated methods rely on finding the maximum over a finite set of candidate points $\textbf{X}^c$ as explained in Section \ref{sec3}. A $(\mu+1)$ evolutionary based algorithm \cite{beyer2001} was used to identify the next sampling point for the remaining methods (see Table \ref{tab1}). Further, \ac{lhs} was used to create the initial $\textbf{X}^0$, candidate $\textbf{X}^c$ and test samples $\textbf{X}^{e}$. The number of initial samples $\minit=10\ndim$ was used as proposed in \cite{loeppky2009}. The same initial and test samples were used for each method in a repetition. The maximum number of samples $\mmax$ was used as stopping criterion. The experimental settings including the dimensions $\ndim$, numbers of initial samples $\minit$, candidate samples $\mcand$, testing samples $\mtest$ and maximum samples $\mmax$ are shown in Table \ref{tab2}.
\begin{table}[!tb]
		\centering
		\begin{minipage}{\columnwidth}
			\caption{Experimental settings depending on the input dimension $\ndim$ of the benchmark functions. $\minit, \mmax, \mcand$ and $\mtest$ represent the initial sample, maximum sample, candidate sample and test sample sizes in this order.}\label{tab2}
			\vspace{\tabspace}
			\centering
			\begin{tabular}{@{}ccccc@{}}
				\toprule
				$\ndim$  & $\minit$ & $\mmax$ & $\mcand$ & $\mtest$\\
				\midrule
				\textbf{1}   & 10  & 40 & 5000   & 100000  \\
				\textbf{2}   & 20  & 140 & 10000 & 100000  \\
				\textbf{3}   & 30  & 180 & 15000 & 100000  \\
				\textbf{4}   & 40  & 250 & 20000 & 100000  \\
				\textbf{6}   & 60  & 250 & 30000 & 100000  \\
				\textbf{8}   & 80  & 250 & 40000 & 100000  \\
				\bottomrule
			\end{tabular}
		\end{minipage}
\end{table}

Emphasis has been placed on comparing adaptive schemes with regard to different aspects such as 
the maximum achieved accuracy after $\mmax$ samples, sample-efficiency and variance of performance. The coefficient of determination $R^2$ \citep{sewall1921} is used as a metric to evaluate the model accuracy. It gives an upper bounded metric and is preferred for regression tasks compared to other metrics (mean squared error, mean absolute error, etc.) \citep{chicco2021}. $R^2$ can be computed as

\begin{equation}
\begin{aligned}
	R^2(\textbf{y}, \tilde{\textbf{y}})&=1 - \frac{\sum_{i=1}^{m} (y_i-\tilde{y}_i)^2}{\sum_{i=1}^{m}(y_i-\bar{y}_i)^2}, \\ & \text{where} \: \bar{y}=\frac{1}{m}\sum_{i=1}^{m}y_i.
\end{aligned}
\nonumber
\end{equation}

The surrogate model $\hat{f}_\iter$ was evaluated in each iteration $\iter$ to compare different adaptive sampling methods. $R^2$ score was calculated from the true response for the test dataset $\textbf{y}^e$ and the surrogate prediction 
$\tilde{\textbf{y}}^e_{\iter}=\left[\hat{f}_\iter(\textbf{x}_1^e), ...,\hat{f}_\iter(\textbf{x}_{\mtest}^e)\right]^T$, with $\textbf{x}^e \in \textbf{X}^e$ and $\hat{f}_\iter$ is the surrogate model trained in the $\iter$-th iteration from $\Set{D}_\iter$.

Additionally, we propose $R^2\sub{Area}\colon \R^{s+1} \rightarrow \R^{[0,1]}$ as a novel metric to measure the overall performance and sample-efficiency of the adaptive methods over $s=(\mmax-\minit-1)$ intervals. $R^2\sub{Area}$ can be computed as
\begin{equation}
	R^2\sub{Area}(\textbf{r}^2)= \frac{f_{s}(\textbf{r}^2) + f_{tr}(\textbf{r}^2)}{s} \label{eq37}
\end{equation}
where $f\sub{s}$ is the composite Simpson rule \citep{venkateshan2014}
\begin{equation}
	\hspace*{-0.2cm}f\sub{s}(\textbf{r}^2)=\frac{1}{3} \left( r^2_{\xi(s)} \hspace*{-0.1cm}+ 2 \sum_{k=1}^{s'-1} r^2_{2k+\xi\sub(s)} \hspace*{-0.1cm}+ r^2_{s} + 4 \sum_{k=1}^{s'} r^2_{2k-1+\xi(s)} \right)\nonumber
\end{equation}
with $s'=(s-\xi(s))/2$, $\textbf{r}^2= \left[ r^2_{0}, ..., r^2_{s} \right]^T $ and $r^2_i = \max\left(0, R^2(\textbf{y}^e, \tilde{\textbf{y}}^e_i)\right)$.
The parity indicator function $\xi(\cdot)$ is defined as
\begin{equation}
	\xi(s)=
	\begin{cases}
		0, & \text{if} \ (-1)^{s}=1, \\
		1, & \text{else}.
	\end{cases}
\nonumber
\end{equation}
If $s$ is odd, then the trapezoidal rule
\begin{equation}
f\sub{tr}(\textbf{r}^2)=\frac{r^2_{0} + r^2_{1}}{2}\xi(s)\nonumber
\end{equation}
 is applied to the first interval.
$R^2\sub{Area}$ allows quantifying the entire sampling history into an easy to interpret scalar value.

As surrogate model, $\ac{gp}(\mu(\mathbf{x}), k(\mathbf{x}, \mathbf{x}', l))$ with zero mean function $\mu(\mathbf{x})=0$, $k(\cdot)$  as the Mat\'ern kernel ($\nu=3/2$), and $\sigma^2_{f}=1$ was used to generate the presented results in the following. 
The output $y$ of the surrogate model were normalized during training and prediction: $y\sub{norm}=(y-y_\mu)/y_\sigma$, where $y_\mu$ and $y_\sigma$ are the mean and standard deviation, in this order, calculated from the training data.
The normalization was reversed for the estimation of the test metrics.
The adjustment factor $\alpha\wmmse=1$ was used for \ac{wmmse}. The committee in \ac{masa} consisted of five \acp{gp} with different covariance functions: squared exponential, two Mat\'ern functions ($\nu = 3/2$ and $\nu=5/2$), dot-product function and a rational quadratic function. 
A \ac{gp} model with posterior mean function $\hat{f}_{\eloo}$ with Mat\'ern kernel ($\nu=3/2$) was used to approximate \ac{loocv} error in Eq. \eqref{eq_aq_dlased}. Moreover, Delaunay triangulation \citep{lee1980} was used to select the support points in Eq. \eqref{eq33}. 
The $m\sub{edges}=2^{\ndim}$ edges of the design space were sampled after the initial samples in order to calculate the exploration component in Eq.~\eqref{eq33} for all $\textbf{x} \in \VSpace{X}$ and to use \ac{dlased}. The acquisition function in Eq.~\eqref{eq_aq_dlased} was maximized for the following samples. 
Further implementation details can be found in \ref{secB}.

\subsection{Results for 2 - 4D Benchmark Functions}\label{sec4_2}
\begin{figure*}[!htb]
	\centering
	\includegraphics[width=1.0\textwidth]{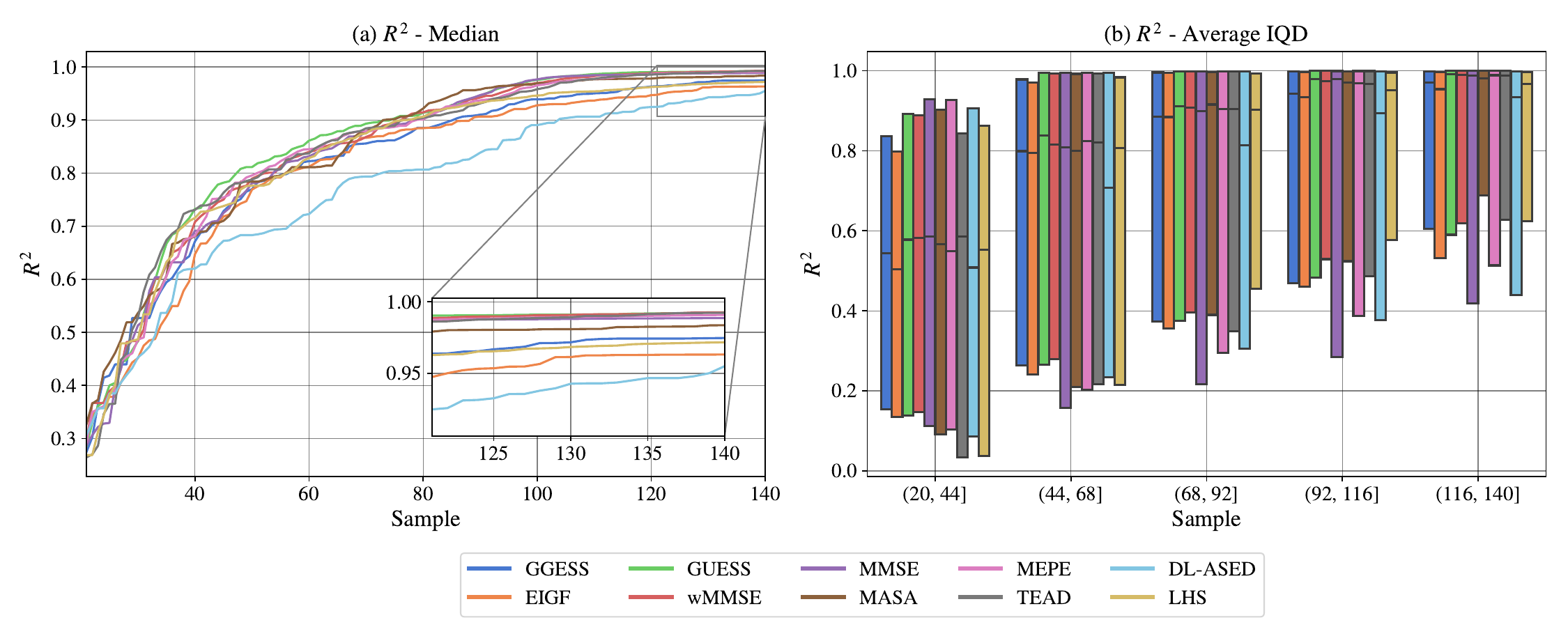}
	\caption{Comparison of different adaptive sampling strategies with \ac{gp} surrogate model across 10 repetitions for 2D benchmark functions: $\Bbeker$, $\Begg$, $\Bhimm$, $\Bbran$, $\Bdrop$, \Bmicha{2} and \Bschw{2} \change{(see \ref{secC})}. (a) $R^2$ median over the conducted benchmarks and repetitions, (b) \ac{iqd} of the average $R^2$ within each group; boxes indicate the 25\% and 75\% quantile and the horizontal line the median.}\label{fig1}
\end{figure*}
One figure with two different plots is presented for each input dimension of the benchmark functions. The left plot shows the $R^2$ score over the sample number, 
where the median of the curves resulting from 10 runs on each benchmark function is displayed for each strategy. Rolling max was performed, based on the assumption that the best model from previous iterations is stored and can be used in the case when no subsequent models could improve the accuracy. The \acf{iqd} of the means is displayed for each group of samples on the right plot. The bin size of the grouped samples was selected to be equal for each of the five groups used throughout.

Figure \ref{fig1} presents the aggregated results for the 2D benchmark functions. It can be seen that the median model performance does not improve much more after $140$ samples.
For the \Bschw{2} and $\Begg$ functions, the model accuracy did not converge to a value close to $1$ within $\mmax$ samples due to the high non-linearity and multimodality of these responses. The variance decreased with increasing number of samples on average for all tested methods (Figure \ref{fig1}b). Using \ac{wmmse} yielded the highest $R^2$ score, closely followed by \ac{guess}, \ac{tead} and \ac{mepe}. The model based on \ac{dlased} achieved the lowest median score. \ac{lhs} and most adaptive strategies showed similar speed of accuracy improvement within the first 100 samples. Later, \ac{guess}, \ac{wmmse}, \ac{mepe}, \ac{mmse} and \ac{tead} overperformed \ac{lhs} and reach on median a $R^2$ score close to $0.99$, as opposed to $R^2\approx0.97$ for \ac{lhs}. \ac{guess} showed significant improvement over \ac{ggess} regarding the overall performance based on the $R^2$ score. 

\begin{figure*}[!htb]%
	\centering
	\includegraphics[width=1.0\textwidth]{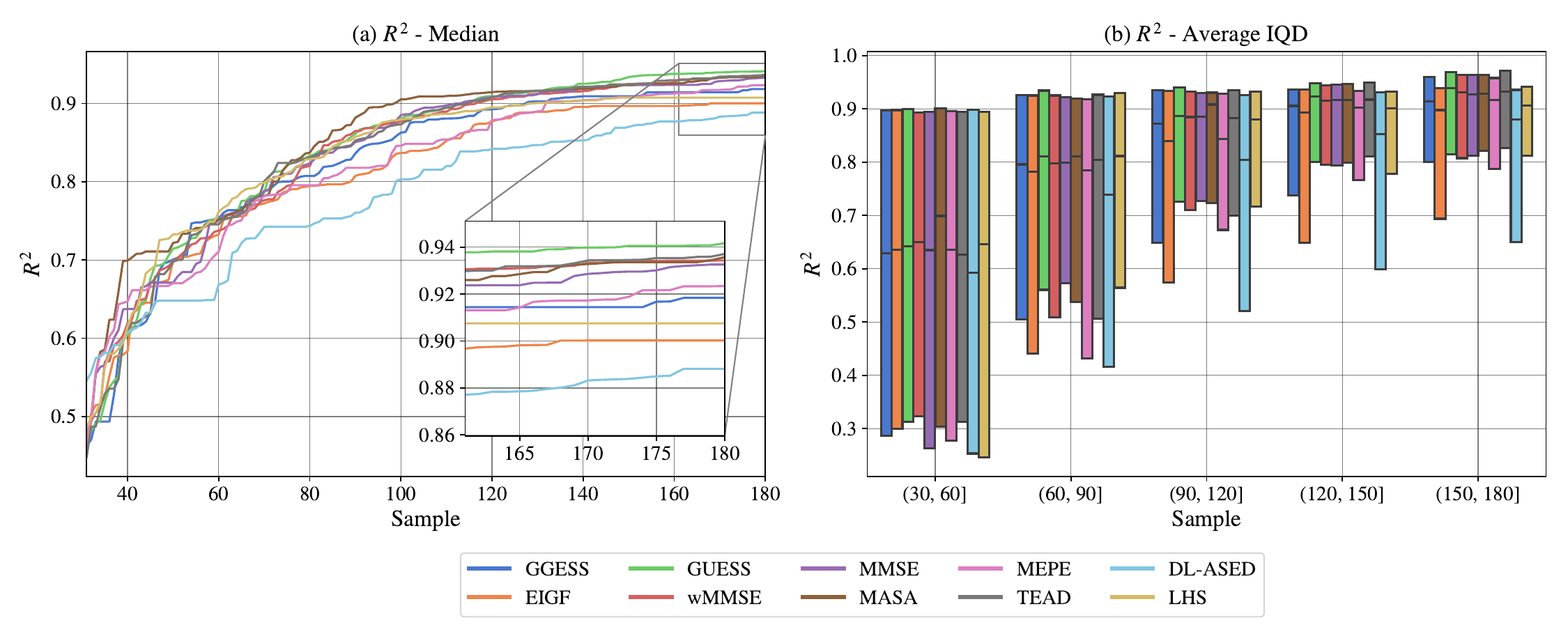}
	\caption{Comparison of different adaptive sampling strategies with \ac{gp} surrogate model across 10 repetitions for 3D benchmark functions: \Back{3}, \Brose{3}, \Bmicha{3} and $\Bishi$ \change{ (see \ref{secC})}. (a) $R^2$ median over the conducted benchmarks and repetitions, (b) \ac{iqd} of the average $R^2$ within each group; boxes indicate the 25\% and 75\% quantile and the horizontal line the median.}\label{fig2}
\end{figure*}

The 3D results are shown in Figure \ref{fig2}, where it is visible that the methods improved not much more after $140$ samples. All adaptive strategies except for \ac{eigf} and \ac{dlased} outperformed the baseline \ac{lhs} in this set of functions. However, \ac{lhs} showed good performance in the beginning until $100$ samples. The purely exploration based \ac{mmse} also worked well with the 3D benchmark functions. Overall, \ac{guess} reached the highest median $R^2$ score, while \ac{dlased} showed the worst median results. 

\begin{figure*}[!htb]
	\centering
	\includegraphics[width=1.0\textwidth]{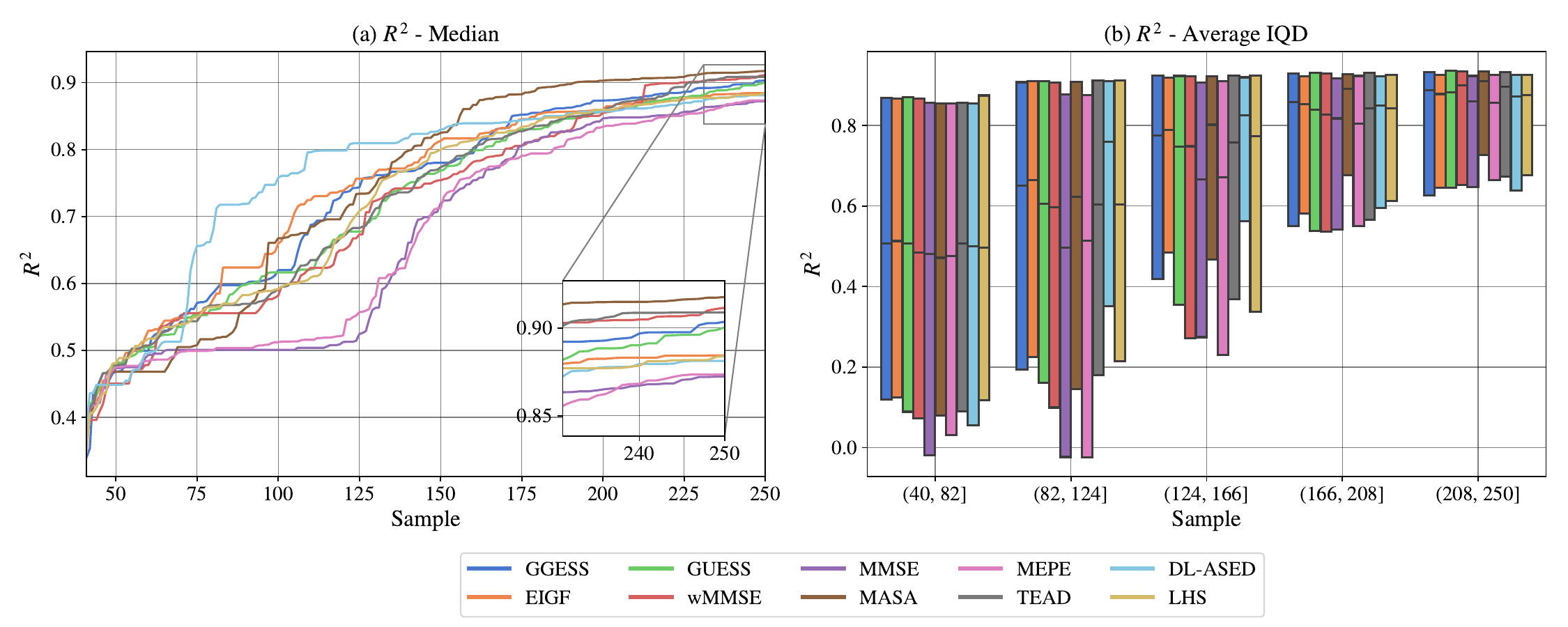}
	\caption{Comparison of different adaptive sampling strategies with \ac{gp} surrogate model across 10 repetitions for 4D benchmark functions: \Back{4}, \Brose{4}, \Bmicha{4} and \Bstyb{4} \change{(see \ref{secC})}. (a) $R^2$ median over the conducted benchmarks and repetitions, (b) \ac{iqd} of the average $R^2$ within each group; boxes indicate the 25\% and 75\% quantile and the horizontal line the median.}\label{fig3}
\end{figure*}

Figure \ref{fig3} shows the 4D benchmark results. In contrast to the previously seen behavior in Figure \ref{fig1} and Figure \ref{fig2}, \ac{dlased} showed a rapid performance increase between 75 and 125 samples compared to the other methods. This can be explained due to the sampling of the edges described in Section \ref{sec4_1}, since the \Brose{4} and \Bstyb{4} functions have large variances at the edges of the design domain. Although beneficial for some of the investigated functions, the sampling of all edges can become a burden for higher-dimensional problems and requires exponentially more samples as given before (see \eg Figure \ref{figA2}). \ac{masa} achieved the highest median $R^2$ score followed by \ac{wmmse} and \ac{tead} for the set of 4D benchmark functions.

\subsection{Results for 1 - 8D Benchmark Functions}\label{sec4_3}
The adaptive sampling strategies were compared across all conducted benchmarks to estimate the overall performance. 
Results for 6 and 8 dimensions are given in \ref{secA_1_1}. The 1D benchmark functions with $\Bgram$, $\Bhsingle$ and $\Bhtwo$ are not explicitly displayed.
Ranks 1 to 10 (smaller is better) were assigned based on the highest $R^2$ and $R^2\sub{Area}$ score achieved in a run for each repetition. The mean rank and standard error of the mean are displayed for both scores in Table \ref{tab4}. 
Additionally, the mean and standard error for the $R^2$ and $R^2\sub{Area}$ score were calculated based on the highest metric score in each repetition of the benchmark (Table \ref{tab4}).
\begin{table}[!tb]
		\centering
		\begin{minipage}{\columnwidth}
			\caption{Average $R^2$, $R^2\sub{Area}$ and respective ranks together with the standard errors for all conducted benchmarks, 10 repetitions and \ac{gp} surrogate model (including results from \ref{secA_1_1}). Average rank is calculated from the highest $R^2$ score achieved in a run for each repetition. Bold numbers represent the best, and underlined the second best result.}\label{tab4}
			\vspace{\tabspace}
			\centering
			\begin{tabular}{@{}ccccccccc@{}}
				\toprule
				\textbf{Method}  & $\mathbf{R^2}$& $\mathbf{R^2}$ \textbf{SE} & $\mathbf{R^2_{Area}}$  &  $\mathbf{R^2_{Area}}$ \textbf{SE} & \textbf{Rank} $\mathbf{R^2}$& \textbf{Rank} $\mathbf{R^2}$ \textbf{SE} & \textbf{Rank} $\mathbf{R^2_{Area}}$& \textbf{Rank} $\mathbf{R^2_{Area}}$ \textbf{SE} \\
				\midrule
				\textbf{\ac{ggess} } &              0.758 &              0.020 &              0.663 &                0.020 &              5.742 &                0.156 &                  5.658 &                     0.163 \\
				\textbf{\ac{eigf}  } &              0.744 &              0.020 &              0.657 &                0.020 &              6.762 &                0.147 &                  6.369 &                     0.157 \\
				\textbf{\ac{guess} } &              0.768 &              0.020 &     \textbf{0.675} &                0.020 &  \underline{3.785} &                0.146 &         \textbf{4.065} &         \underline{0.141} \\
				\textbf{\ac{wmmse} } &              0.765 &              0.020 &              0.671 &                0.020 &              4.027 &    \underline{0.142} &                  4.581 &                     0.159 \\
				\textbf{\ac{mmse}  } &              0.733 &              0.020 &              0.638 &                0.021 &              6.665 &                0.176 &                  6.973 &                     0.177 \\
				\textbf{\ac{masa}  } &  \underline{0.771} &              0.019 &  \underline{0.674} &    \underline{0.020} &              4.988 &                0.178 &                  4.473 &                     0.153 \\
				\textbf{\ac{mepe}  } &              0.752 &              0.020 &              0.657 &                0.020 &              5.531 &                0.166 &                  5.973 &                     0.176 \\
				\textbf{\ac{tead}  } &     \textbf{0.773} &  \underline{0.019} &              0.672 &                0.020 &     \textbf{3.350} &       \textbf{0.115} &      \underline{4.269} &            \textbf{0.140} \\
				\textbf{\ac{dlased}} &              0.718 &              0.021 &              0.632 &                0.021 &              7.738 &                0.149 &                  7.162 &                     0.193 \\
				\textbf{\ac{lhs}   } &              0.758 &     \textbf{0.019} &              0.665 &       \textbf{0.020} &              6.254 &                0.171 &                  5.308 &                     0.177 \\
				\bottomrule
			\end{tabular}
		\end{minipage}
\end{table}
The results show that \ac{guess} achieved the highest $R^2\sub{Area}$ score with its respective rank, the second  best average rank over all tested cases based on $R^2$, and the third highest score based on the average $R^2$. \ac{tead} reached the highest $R^2$ score and rank, together with the second best $R^2\sub{Area}$ rank on average. Furthermore, \ac{masa} achieved the second highest $R^2$ and $R^2\sub{Area}$ on average. The competitive advantage of \ac{masa} is the access to multiple models with different covariance functions; as opposed to the other adaptive strategies which are restricted to a \ac{gp} with Matérn kernel. 
Nevertheless, no single sampling strategy dominated its competitors across all functions. Moreover, it was observed that nearly each investigated method (except for \ac{dlased} and \ac{eigf}) achieved the best position in at least one of the benchmark functions\textsuperscript{\ref{github}}. 
Which sampling strategy achieved the highest score overall was problem dependent. For example, \ac{mmse} achieved good results for 1 and 2-dimensional problems, while performing worse in higher dimensions compared to its competitors. In contrast, \ac{ggess} was found to perform better for higher-dimensional problems. 
We note that although \ac{tead} achieved the highest $R^2$ score and the respective rank, \ac{guess} provided overall the best performance with less iterations as indicated by $R^2\sub{Area}$ and achieved on average $R^2$ scores close to the best-performing strategies. It was found that \ac{lhs} overperformed nearly half of the adaptive strategies on average regarding $R^2$ and $R^2\sub{Area}$.  Most of the strategies considering both exploration and exploitation showed on average improved performance over the purely exploration based \ac{mmse}. 

\section{Conclusion}\label{sec5}
This work proposed a novel adaptive sampling strategy called \ac{guess} and compared it to some of the most recent adaptive sampling strategies to improve the global model accuracy within a unified framework using various benchmark functions. 
The proposed acquisition function guiding the sampling process leverages the predicted standard devia\-tion of the surrogate model to both balance the exploitation term based on a approximation of the second and higher-order Taylor expansion values and explore new regions with high predictive variance. The proposed method can be used with any probabilistic model with a heteroscedastic variance estimate and was tested for single-response problems. A straightforward extension to handle multi-response problems could be to maximize a sum of the acquisition function values of each model output. Testing this approach is left for future research.

Due to the low number of comparisons, a large-scale comparative study is conducted to investigate the differences in recent developments. The presented methods were compared using a \ac{gp} and 1 to 8-dimensional deterministic bench\-mark functions. 
\ac{guess} achieved on average the highest sample efficiency regarding model accuracy compared to 9 other sampling strategies, and the second-highest accuracy overall, based on ranking all experiments. No single sampling strategy dominated its competitors across all functions. Never\-theless, the gradient-based methods \ac{guess} and \ac{tead}, as well as the committee-based method \ac{masa}, achieved the best performance within the conducted study on average. 

Finally, our ablation study shows that the choice of surrogate model can have a great influence on the individual method performance and is therefore an important factor to consider. However, \ac{masa}, along with \ac{guess}, provided the best results across all tested models on average. Additionally, we found using a more suitable surrogate model can have a greater impact on the achieved accuracy, and thus the sample efficiency compared to the choice of sampling strategy. In this context, a more suitable model is expected to achieve comparatively higher accuracy, when trained on the same data set as a less suitable model. However, the decision of the most suitable model for the task at hand remains a non-trivial question. Thus, further emphasis should be placed on exploring the influence of the surrogate model choice on adaptive sampling strategies. 
Moreover, determining the number of initial samples and selecting the stopping criteria are important yet open questions, as only a limited number of studies have addressed these issues.

\section*{Author statements}
\textbf{Sven Lämmle}: Conceptualization, Methodology, Implementation, Writing - Original Draft. 

\textbf{Can Bogoclu}: Conceptualization, Methodology, Writing - Review \& Editing. 

\textbf{Kevin Cremanns}: Methodology, Writing - Review. 

\textbf{Dirk Roos}: Supervision, Writing - Review.

%\section*{Declaration of competing interest}
%The authors declare that they have no known competing financial interests or personal relationships that could have appeared to influence the work reported in this paper. 

\section*{Acknowledgment}
This work was supported by ZF Friedrichshafen AG.

\bibliographystyle{elsarticle-num} 
\bibliography{Library_Paper_BibTex}

\appendix
\newpage

\section{Complementary Results}\label{secA}

\subsection{\change{Influence of increasing Dimensionality on Method Performance}}\label{secA_1}
\change{We present the results regarding influence of increasing dimensionality on performance of adaptive sampling strategies in the following. Therefore, we compare the methods for 6 and 8-dimensional benchmark functions and study the behavior of \ac{guess} up to 32 dimensions. Furthermore, we discuss challenges and approaches for adaptive sampling to overcome the curse of dimensionality.}
\subsubsection{Results for 6 - 8D Benchmark Functions}\label{secA_1_1}
\begin{figure*}[!b]%
	\centering
	\includegraphics[width=1.0\textwidth]{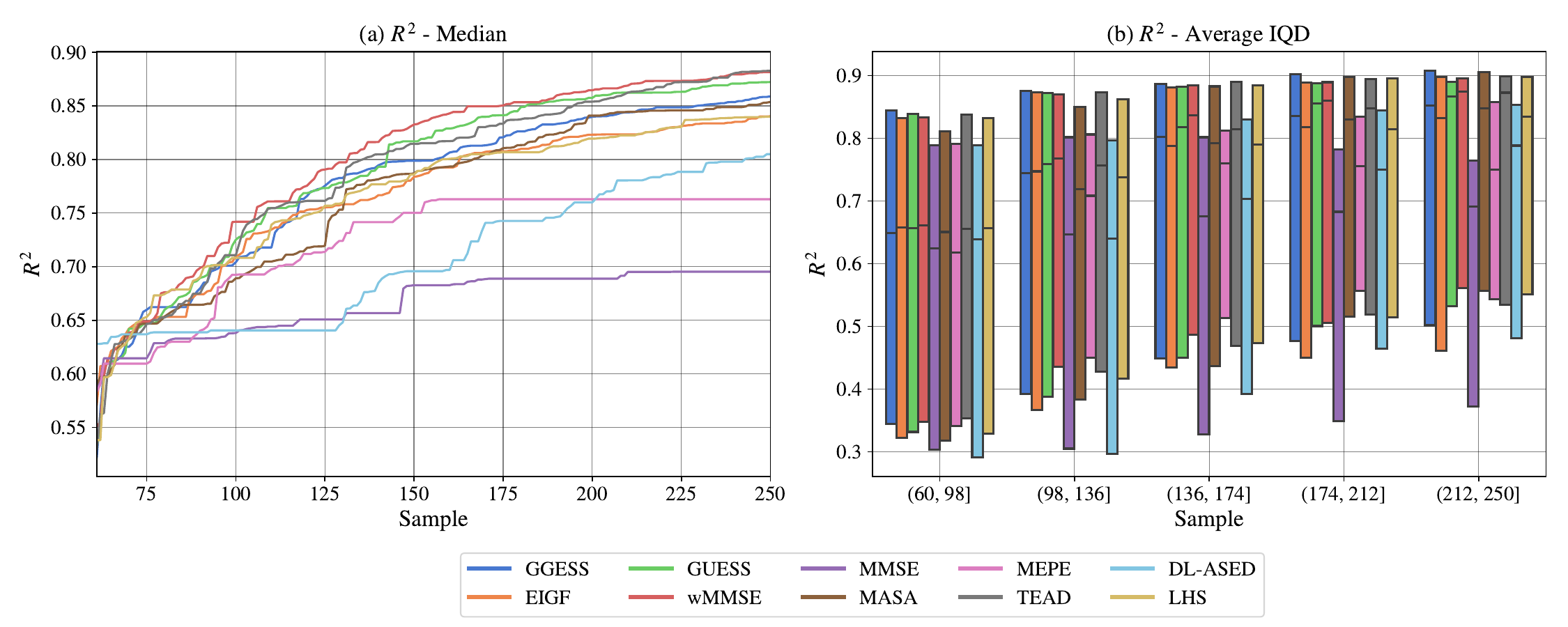}
	\caption{Comparison of different adaptive sampling strategies with \ac{gp} surrogate model across 10 repetitions for 6D benchmark functions: \Back{6}, \Brose{6}, \Bmicha{6} and $\Bhart$ \change{ (see \ref{secC})}. (a) $R^2$ median over the conducted benchmarks and repetitions, (b) \ac{iqd} of the average $R^2$ within each group; boxes indicate the 25\% and 75\% quantile and the horizontal line the median.}\label{figA1}
\end{figure*}

\begin{figure*}[!tb]%
	\centering
	\includegraphics[width=1.0\textwidth]{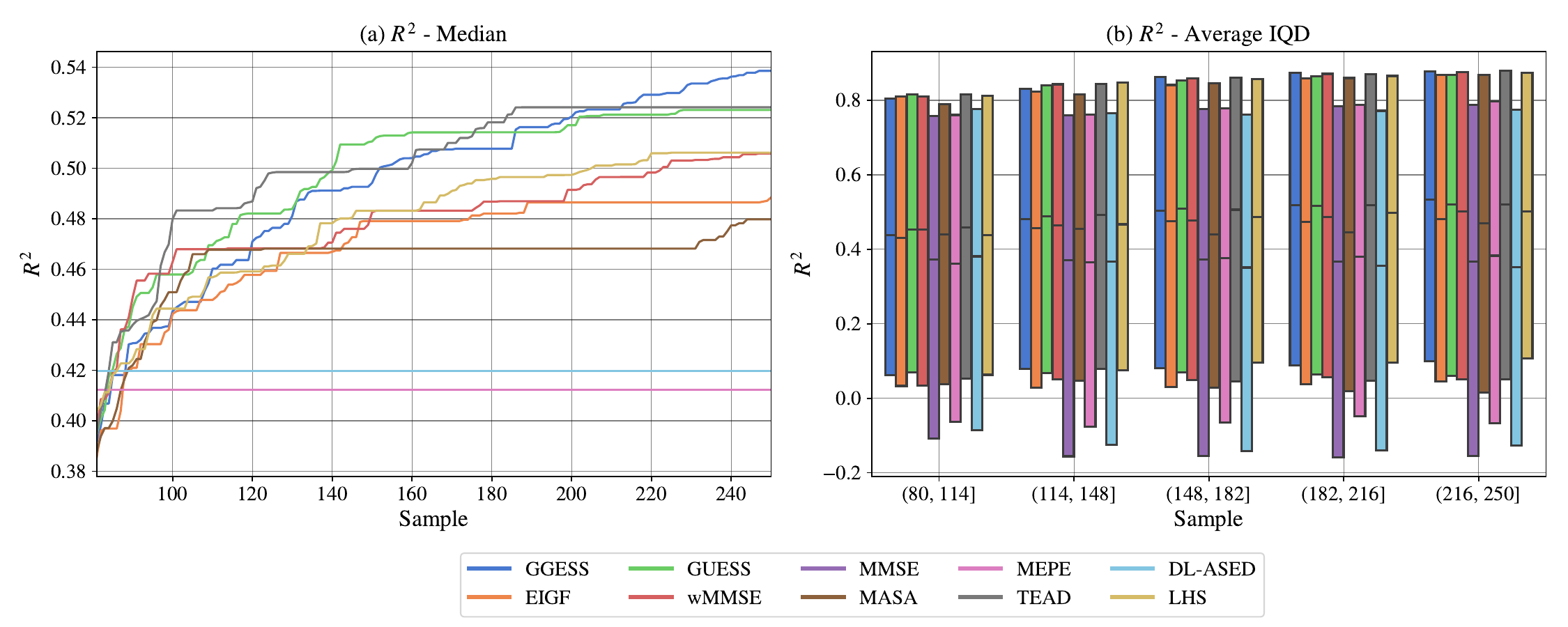}
	\caption{Comparison of different adaptive sampling strategies with \ac{gp} surrogate model across 10 repetitions for 8D benchmark functions: \Back{8}, \Brose{8}, \Bmicha{8} and \Bstyb{8} \change{(see \ref{secC})}. (a) $R^2$ median over the conducted benchmarks and repetitions, (b) \ac{iqd} of the average $R^2$ within each group; boxes indicate the 25\% and 75\% quantile and the horizontal line the median.}\label{figA2}
\end{figure*}

6 and 8-dimensional benchmarks were conducted to study the influence of increasing dimensionality with constant sample size. Results of the 6D benchmark are shown in Figure \ref{figA1}. \ac{tead} achieved the highest accuracy closely followed by \ac{wmmse} and \ac{guess}.  \ac{mmse} underperformed compared to other algorithms. 8D benchmarks results are depicted in Figure \ref{figA2}. It can be seen that gradient-based strategies outperformed most non-gradient based methods in the later half and achieved the highest median $R^2$ score in comparison. 
Nevertheless, \ac{lhs} showed comparable performance to all methods and even an overall improvement over non-gradient based strategies regarding the median $R^2$. 
This is not very surprising, since the initial small surrogate model accuracy could lead to erroneous information for the acquisition functions and, thus, misguide the sampling procedure. \ac{mepe}, \ac{mmse} and \ac{dlased} did not achieve an improvement after $250$ samples compared to the initial data set. The exhaustive sampling of $2^8$ edges could be the reason for \ac{dlased}, which lead to not using the acquisition function. Additionally, it could be observed that the variance increased for most methods with newly added samples (\ref{figA2}b).
This contradicts the observed behavior in Section \ref{sec4_2}, where the variance decreased with increasing number of samples. One explanation could be the faster improvement of accuracy for \Back{8} and \Brose{8} functions relative to \Bmicha{8} and \Bstyb{8}, leading to an increase in variance.
\subsubsection{\change{Behavior of \ac{guess} in Higher Dimensions}}\label{secA_1_3}
\change{We study the behavior of \ac{guess} for higher-dimensional problems up to 32 dimensions and compare it to the baseline \ac{lhs}. \ac{svgp} is used to carry out the benchmarks for dimensions greater than 8. \ac{svgp} can mitigate the computational burden and allow for much larger datasets. Instead of conditioning on all available samples, an \ac{svgp} model learns a selected subset of so called \emph{inducing points}. This can reduce time complexity from $\mathcal{O}(m^3)$ for \ac{gp} to $\mathcal{O}(m\sub{u}^3)$ since generally $m\sub{u}<<m$, where $m\sub{u}$ are the number of \emph{inducing variables} $\textbf{u}\in \VSpace{Y}$, defined at \emph{inducing locations} $\textbf{z} \in \VSpace{X}$. This already indicates that the computational effort will be constant after observing $m \geq m\sub{u}$ samples, since $m\sub{u}$ is constant.}
	
\change{The variational approach defines an approximate posterior over the inducing variables as $q(\textbf{u})=\mathcal{N}(\boldsymbol{\mu}_{U}, \Sigma_U)$, where mean vector $\boldsymbol{\mu}_{U}\in\R^{m\sub{u}}$ and covariance matrix $\Sigma_U\in\R^{m\sub{u}\times m\sub{u}}$ are variational parameters that have to be inferred. Therefore, inducing locations, variational and \ac{gp} parameters are learned jointly by maximizing a lower bound to the marginal likelihood called evidence lower bound \citep{hensman2013}.}

\change{The approximate posterior mean and variance is given by
\begin{align*}
	\hat{f}\sub{SGP}( \textbf{x}^* ; \boldsymbol{\theta}\sub{GP}) = \textbf{k}\sub{Z}^{T} \textbf{K}\sub{Z}^{-1} \boldsymbol{\mu}_{U} %\label{eq_sgp_mu}
\end{align*}
\begin{align*}
	\mathbb{V}\left[\hat{f}\sub{SGP}(\textbf{x}^*; \boldsymbol{\theta}\sub{GP})\right] = \textbf{K} + \textbf{k}\sub{Z}^T \textbf{K}\sub{Z}^{-1}(\Sigma_U-\textbf{K}\sub{Z})\textbf{K}\sub{Z}^{-1}\textbf{k}\sub{Z}  %\label{eq_sgp_var}
\end{align*}
where $\textbf{k}\sub{Z}$ and $\textbf{K}\sub{Z}$ are defined similarly to $\textbf{k}$ and  $\textbf{K}$ by using inducing locations $\textbf{z}$ instead of $\textbf{x}$, respectively. See \citep{hensman2013, burt2020} for more details.}

\begin{figure*}[!tb]%
	\centering
	\includegraphics[width=1.0\textwidth]{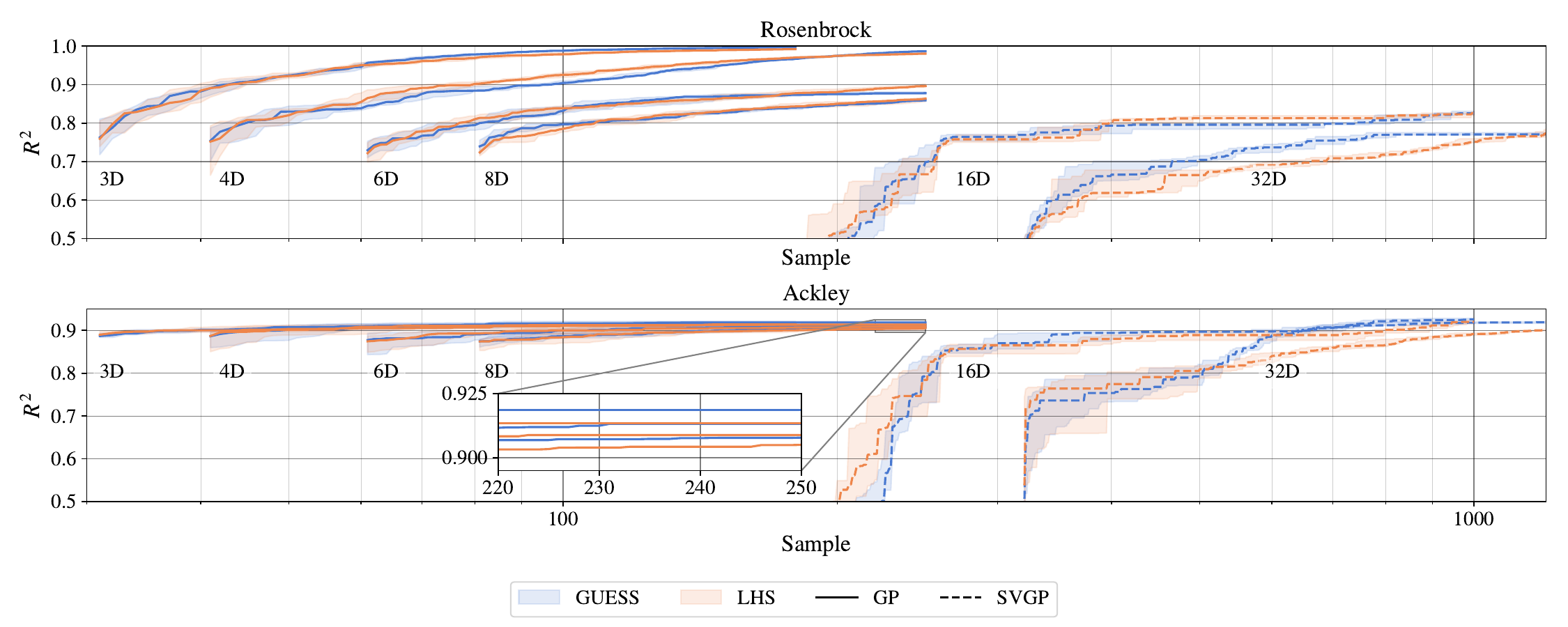}
	\caption{\change{Comparison of \ac{guess} and \ac{lhs} for 3 to 32 dimensions across 10 repetitions of \Brose[]{} (top) and \Back[]{} (bottom) benchmark functions (see \ref{secC}). \ac{gp} was used for 3 to 8 dimensions and \ac{svgp} for the remaining. \ac{iqd} of the $R^2$ score across proposed samples is displayed; filled area indicate the 25\% and 75\% quantile and line the median. Magnified area for \Back[]{} shows 4D, 6D and 8D from top to bottom, respectively for \ac{guess} and \ac{lhs}.}}\label{fig_highdim}
\end{figure*}

\change{Figure \ref{fig_highdim} shows the influence of increasing dimensionality on method performance. Similar experimental settings to Section \ref{sec4_1} and \ref{secA_1_1} were used. For \ac{svgp}, we used $m\sub{u}=256$ and Matérn kernel ($\nu=3/2$). Above 8 dimensions, $m\sub{cand}=80000$ were used and the model was trained only every second iteration. Even if the model was not trained at an iteration, new samples are still added to the model but the same kernel parameters as the previous iteration were used.
Considering only results from 16 and 32 dimensions, \ac{guess} achieved the highest median $R^2$ ($0.86$) and $R\sub{Area}^2$ score ($0.806$) on average, in contrast to \ac{lhs} with $R^2= 0.855$ and $R\sub{Area}^2=0.792$. It can be observed that with growing dimensionality, the sample size increases to achieve an equivalent $R^2$ score (see Table \ref{tab_acc_highdim}).}
\begin{table}[!tb]
	\centering
	\begin{minipage}{\columnwidth}
		\caption{\change{Average number of samples needed to achieve a defined accuracy target ($R^2$) across dimensions based on results from \ac{guess} and \ac{lhs}. \ac{gp} was used for 3 to 8 dimensions and \ac{svgp} for the remaining. None of the runs could achieve the accuracy goal within 1200 samples for \Brose[]{} and 32D. Runs underperforming the accuracy target are not considered.}}\label{tab_acc_highdim}
		\vspace{\tabspace}
		\centering
		\begin{tabular}{@{}cc|cccccc@{}}
			\toprule
			Function & Target $R^2$ & 3D  & 4D & 6D & 8D & 16D & 32D\\
			\midrule
			\Brose[]{} & $\geq0.8$ & 33  & 46 & 80 & 107 & 547 & - \\
			\Back[]{} & $\geq0.9$ & 45 & 57 & 103 & 160 & 682 & 844 \\
			\bottomrule
		\end{tabular}
	\end{minipage}
\end{table}
\subsubsection{\change{Challenges and Approaches in High Dimensions}}\label{secA_challenges}
\change{Adaptive sampling in high dimensions is challenging, yet such problems are common in different real-world applications. The curse of dimensionality concerns both the surrogate model and the optimization of the acquisition function. With increasing dimensionality and sample requirements to achieve sufficient accuracy \ac{gp} may become infeasible due to $\mathcal{O}(\nsamp^3)$ time and $\mathcal{O}(\nsamp^2)$ memory complexity \citep{rasmussen2006}, therefore sparse approximations \citep{hensman2013}, or alternatives like batch \acp{dgcn} \citep{cremanns2021} or \acp{pnn} have to be used. Optimizing Eq. \eqref{eq17} becomes also increasingly difficult, especially for candidate based acquisition functions, since larger candidate sets have to be used to cover the whole optimization domain. Previous research \citep{chen2021} introduced a partitioning of the optimization domain based on Voronoi cells to identify sampling regions, aiming to improve the optimization efficiency which does not account for the modeling difficulties.}

\change{In the related field of \ac{bo}, different approaches dealing with high dimensionality were proposed \citep{wang2023}. However, most assume the high-dimensional objective function has lower intrinsic dimensionality, \ie the response function is insensitive to changes in most of the dimensions. Therefore, global sensitivity analysis \citep{saltelli2007} could be used to select only input variables which have important contribution to the response variability. Alternatively, embeddings based on linear \citep{wang2016, shlens2014} or non-linear projections \citep{scholkopf1997, rumelhart1987} could be used to learn a reduced latent representation of the input. However, none of these methods are efficient if the response function is sensitive to all dimensions. Although there is some ongoing effort to solve them \citep{wang2023}, such problems remain important opportunities for future research.}

\subsection{Influence of Surrogate Model}\label{secA_models}
The adaptive sampling strategies were tested with the 4D benchmark functions using \ac{dgcn} and \ac{pnn} as an alternative to \ac{gp}. The aim of this study was to investigate the influence of the model choice. 

\subsubsection{DGCN}\label{secA_dgcn}
\ac{dgcn} is an anisotropic and non-stationary \ac{gp}, where each input variable has its own length scale $l$ and noise variance prediction $\sigma\sub{n}^2$.
$l$ and $\sigma\sub{n}^2$ are learned by an \ac{ann} such that for an input $\mathbf{x}\in\VSpace{X}$, the \ac{gp} parameters can be predicted
$\hat{f}\sub{ANN}: \VSpace{X} \times \VSpace{X} \rightarrow \R^{n_{\theta}}$, where $n_{\theta}$ are the number of \ac{gp} parameters. Since the length scale is no longer fixed over the training samples, the covariance function has to be reparameterized using two length scales $l$ and $l'$. For the Mat\'ern kernel this can be done by replacing $r/l$ in Eq.~\eqref{eq7} with $\sqrt{\sum_{i=1}^{n} \left( x_i / l - x'_i / l' \right)^2}$.

Five different covariance functions are combined to approximate complex functions; squared exponential, absolute exponential, two Mat\'ern functions ($\nu = 3/2$ and $\nu=5/2$) and a rational quadratic function. The free parameters $\textbf{l}$ of the kernel functions are then learned by the \ac{ann}.
Stochastic gradient descent (\eg\citep{kingma2015}) is used to obtain \ac{ann} parameters by maximizing the marginal log-likelihood
\begin{align*}
	\hspace*{-0.2cm}\log p(\textbf{y} \vert \textbf{X}, \hat{f}\sub{ANN}) = 
	- \dfrac{1}{2}\textbf{y}\textbf{K}\sub{C}^{-1}\textbf{y} - \dfrac{1}{2} \log \vert \textbf{K}\sub{C} \vert - \dfrac{m}{2} \log(2\pi) 
\end{align*}
where the covariance matrix $\textbf{K}\sub{N}$ in Eq.~\eqref{eq8} is replaced with
\begin{equation*}
	\textbf{K}\sub{C} = \sum_{i=1}^{n\sub{C}} K_i(\textbf{X}, \textbf{X};\hat{f}\sub{ANN}(\textbf{X})) + \sigma\sub{n}^2(\textbf{X}) \textbf{I}
\end{equation*}
where $K \colon \VSpace{X}^{m} \times \VSpace{X}^{m} \rightarrow \R^{m \times m}$ is the covariance matrix function,  $n\sub{C}$ representing the number of covariance functions used, and $\sigma\sub{n}^2 \colon \VSpace{X}^{m} \rightarrow \R^{m}$ is obtained by a further \ac{ann} (see \citep{cremanns2021}).
Equations~\eqref{eq9} and \eqref{eq10} can be adapted for prediction at a query $\textbf{x}^*$ as
\begin{equation*}
	\hat{f}\sub{DGCN}(\textbf{x}^*; \hat{f}\sub{ANN}) = \textbf{k}\sub{C}^{T} \textbf{K}\sub{C}^{-1} \textbf{y}
\end{equation*}
\begin{align*}
	\mathbb{V}\left[\hat{f}\sub{DGCN}(\textbf{x}^*; \hat{f}\sub{ANN})\right] =  k\sub{C}(\textbf{x}^*, \textbf{x}^*; \tilde{\textbf{l}}^*, \tilde{\textbf{l}}^*)+\hat{\sigma}\sub{n}^2(\textbf{x}^*)-\textbf{k}\sub{C}^{T} \textbf{K}\sub{C}^{-1}\textbf{k}\sub{C} 
\end{align*}
with the combination of the different kernel functions $k\sub{C}(\textbf{x}^*, \textbf{x}^*; \tilde{\textbf{l}}^*, \tilde{\textbf{l}}^*) = \sum_{i=1}^{n\sub{C}} k_i (\textbf{x}^*, \textbf{x}^*; \tilde{\textbf{l}}^*, \tilde{\textbf{l}}^*)$ 
and the vector of correlations as $\textbf{k}\sub{C} = \left[k\sub{C}(\textbf{x}^*, \textbf{x}_1; \tilde{\textbf{l}}^*, \tilde{\textbf{l}}_{1}), ...,k\sub{C}(\textbf{x}^*, \textbf{x}_m; \tilde{\textbf{l}}^*, \tilde{\textbf{l}}_{m}) \right]^T$. $\hat{\sigma}\sub{n}^2(\textbf{x}^*)$ is the predicted noise and $\tilde{\textbf{l}}$ the vector of predicted length scales from the \ac{ann}. Additional details like the net topology, partial derivatives, etc. can be found in \citep{cremanns2021}.

\begin{figure*}[!tb]%
	\centering
	\includegraphics[width=1.0\textwidth]{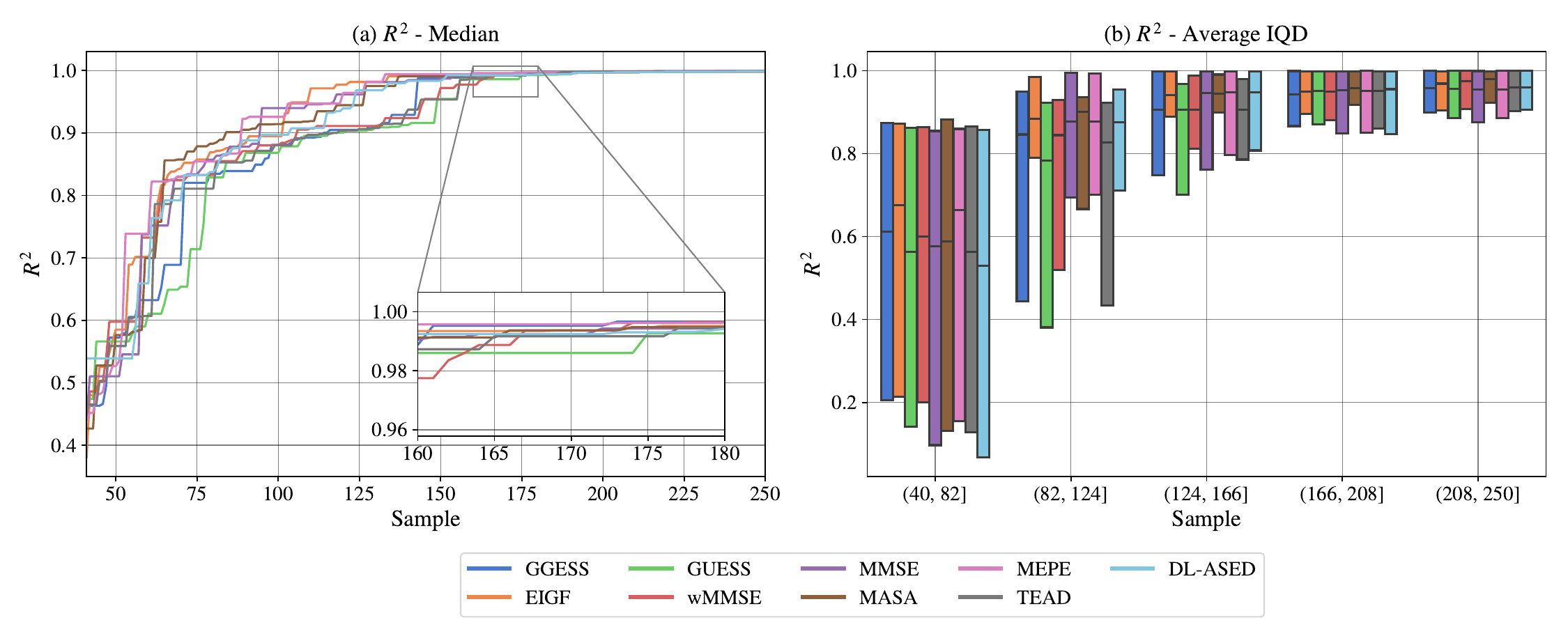}
	\caption{Comparison of different adaptive sampling strategies with \ac{dgcn} surrogate model across 10 repetitions for 4D benchmark functions: \Back{4}, \Brose{4}, \Bmicha{4} and \Bstyb{4}. (a) $R^2$ median over the conducted benchmarks and repetitions, (b) \ac{iqd} of the average $R^2$ within each group; boxes indicate the 25\% and 75\% quantile and the horizontal line the median.}\label{figA4}
\end{figure*}

The results for the 4D benchmark functions are given in Figure \ref{figA4}. \ac{dgcn} achieved a higher accuracy for all sampling strategies compared to \ac{gp} (Figure \ref{fig3}). With \ac{dgcn} and 100 samples, a similar or better median $R^2$ score could be achieved for all methods, compared to adaptive sampling with \ac{gp} and 250 samples.

All methods could achieve a median $R^2$ score close to 1 within 250 samples. It can be seen that \ac{masa} performed robust with both surrogate models, achieving the highest $R^2$ and second highest $R^2\sub{Area}$ score on average with \ac{dgcn}. Regarding the performance, it can be seen that \ac{mmse} and \ac{mepe} performed better with \ac{dgcn}, while the gradient-based acquisition functions got relatively worse\textsuperscript{\ref{github}}. 
This is also reflected within the $R^2\sub{Area}$ ranks, where \ac{ggess}, \ac{tead} and \ac{guess} were the only methods losing ranks compared to \ac{gp}.
This could be partially due to the use of forward differences \cite{dahmen2008} for calculating the gradients of the \ac{dgcn} model. 

We conclude that the impact of the surrogate model choice can be much greater than the adaptive sampling strategy regarding the achieved accuracy, and thus the sample efficiency for the tested case. It can be seen that the choice of surrogate model can change the ranking of the sampling strategies drastically. However, the right choice of adaptive sampling strategy can improve the sample efficiency further, \eg considering the performance difference between \ac{ggess} and \ac{eigf} in Figure \ref{figA4}.

\subsubsection{Other non-kernel Methods}\label{secA_pnn}
Ensembles of bootstrapped \acp{pnn} were investigated alongside the kernel-based methods from the previous section to be used as the surrogate model for adaptive sampling strategies. The \ac{pnn} is an \ac{ann}, where the output neurons parameterize a probability distribution function to capture the aleatoric uncertainty represented in the dataset. A simple feed forward network with parameters $\boldsymbol{\theta}$ and $n_z$ hidden layers can be written as
\begin{equation}
	\begin{aligned}
		&\mathbf{z}_0 = \mathbf{x}, \\  
		&\mathbf{z}_i = \psi_i \left(\mathbf{W}_i \mathbf{z}_{i-1} + \mathbf{b}_i\right) \quad \forall i \in [1, n_z], \\ 
		&\mathbf{y} = \mathbf{z}_{n_z}
	\end{aligned}\label{eq_nn}
\end{equation}
where $\mathbf{W}_i$ are the weights and $\textbf{b}_i$ the biases of the $i$-th layer.  $\mathbf{z}_{i>0}$ represents the resulting latent variable after the linear transformations using $\mathbf{W}_i$, $\textbf{b}_i$ and the non-linear activations $\psi_i(\cdot)$ of the neurons in the $i$-th hidden layer. 
For a Gaussian posterior distribution $\mathcal{N}(\mu, \sigma^2)$ with mean $\mu$ and variance $\sigma^2$, the \ac{pnn} can be compactly written as
\begin{equation}
	\begin{gathered}
		\mu_{\boldsymbol{\theta}}(\textbf{x})  = g^{(1)}_{\boldsymbol{\theta}}(\mathbf{x}), \\
		\sigma_{\boldsymbol{\theta}}(\textbf{x}) = g_{\boldsymbol{\theta}}^{(2)}(\mathbf{x}), \\
		y \sim \mathcal{N}(\mu_{\boldsymbol{\theta}}(\textbf{x}), \sigma^2_{\boldsymbol{\theta}}(\textbf{x}) )
	\end{gathered}\nonumber
\end{equation}
where $g_{\boldsymbol{\theta}}$ represents the network (Eq. \eqref{eq_nn}) with $\mathbf{z}_{n_z} \in \R^2$. The superscripts $(1)$, $(2)$ denote the first and second entry of the output vector.
Furthermore, a loss proportional to the negative log-likelihood is used to train the model parameters $\boldsymbol{\theta}$ \citep{chua2018}
\begin{align}
	\text{loss} (\boldsymbol{\theta}) = \sum_{i=1}^{m} 
	\left(\frac{\mu_{\boldsymbol{\theta}}(\textbf{x}_i) - y_i}{\sigma_{\boldsymbol{\theta}}(\textbf{x}_i)}\right)^2
	+ \log \sigma^2_{\boldsymbol{\theta}}(\textbf{x}_i) \label{eq_pnn_loss}
\end{align}

Ensembles of bootstrapped \acp{pnn} can be used to estimate the epistemic model  uncertainty without introducing new parameters into the network. This makes the training easier compared to a full Bayesian inference \cite{jospin2022}.
The ensemble $\Set{M}\sub{PNN} = \{\hat{f}\sub[, 1]{PNN}, ..., \hat{f}\sub[, n_{en}]{PNN} \}$ can be constructed by sampling $m$ times (with replacement) from $\Set{D}$, creating $n_{en}$ different datasets for each of the $n_{en}$ models. The ensemble is a uniformly-weighted mixture model. Given a \ac{pnn} with Gaussian posterior, the predictive mean and variance of the mixture can be calculated at a point $\textbf{x}^*$ as \citep{lakshminarayanan2017}
\begin{align*}
	\hat{f}\sub{EPNN}( \textbf{x}^* ; \Set{M}\sub{PNN}) = \mu _{Y^*} = \frac{1}{n_{en}} \sum_{j=1}^{n_{en}} \mu_{\boldsymbol{\theta}_j}(\textbf{x}^*) 
\end{align*}
\begin{align*}
	\mathbb{V}\left[\hat{f}\sub{EPNN}(\textbf{x}^*; \Set{M}\sub{PNN})\right] = \sigma_{Y^*}^{2} =  
	\frac{1}{n_{en}} \sum_{j=1}^{n_{en}} \left(\sigma_{\boldsymbol{\theta}_j}^2(\textbf{x}^*) +  \mu_{\boldsymbol{\theta}_j}^2(\textbf{x}^*) \right)  - \mu_{Y^*}^2 
\end{align*}

The advantage of \acp{ann} and \acp{pnn} is the scalability to larger datasets, whereby smaller data can be approximated with \eg non-linear regression in a Bayesian setting. Difficulties can arise with the selection of the net topology together with the right activation functions types, since it is crucial for the model accuracy, albeit non-trivial.  

We tested the 4D benchmark functions with the ensemble \ac{pnn}. Similar experimental settings to Section \ref{sec4_1} and \ref{secA_1_1} were used to investigate the influence of surrogate model. 
The squared error was used for \ac{mepe}, \ac{wmmse} and \ac{dlased}, since \ac{loocv} (Eq.\eqref{14}) can be challenging to compute for \acp{ann} and \acp{pnn}. Further implementation details can be found in \ref{secB}.
\begin{figure*}[!tb]%
	\centering
	\includegraphics[width=1.0\textwidth]{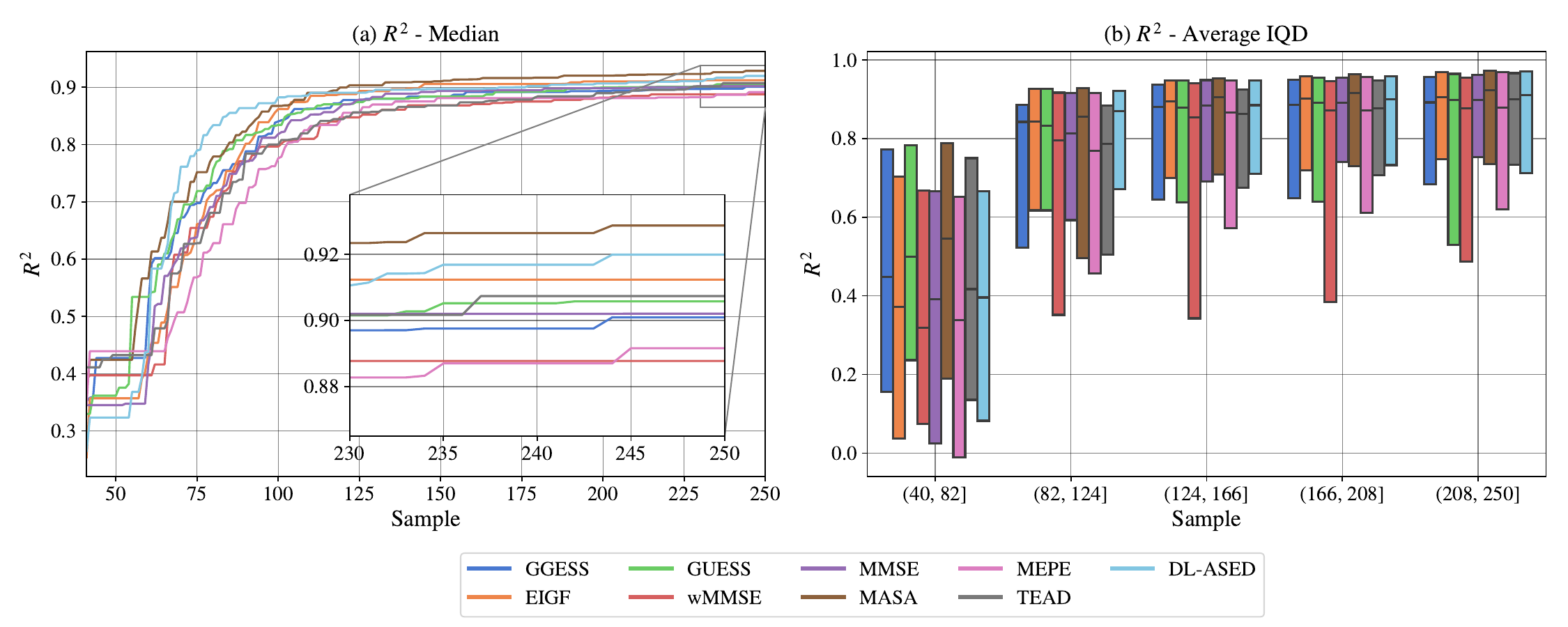}
	\caption{Comparison of different adaptive sampling strategies with \ac{pnn} surrogate model across 10 repetitions for 4D benchmark functions: \Back{4}, \Brose{4}, \Bmicha{4} and \Bstyb{4}. (a) $R^2$ median over the conducted benchmarks and repetitions, (b) \ac{iqd} of the average $R^2$ within each group; boxes indicate the 25\% and 75\% quantile and the horizontal line the median.}\label{figPNN}
\end{figure*}
The results for the 4D benchmark are shown in Figure \ref{figPNN}.
It can be seen that \ac{pnn} achieved on median similar or better results compared to \ac{gp} but performed worse than \ac{dgcn} regarding the $R^2$ score. 

\ac{masa} was found to score on average the highest $R^2\sub{Area}$ value with $0.724$ followed by \ac{ggess} ($0.717$) and \ac{guess} ($0.716$). It can be seen that \ac{wmmse} performed  worse compared to \ac{gp}, which could be due to the replacement of the \ac{loocv}. 
Similar ranks regarding $R^2\sub{Area}$ can be reported for the 4D benchmark functions compared to \ac{gp} for most methods\textsuperscript{\ref{github}}. 
The majority of methods stayed within one rank difference, while \ac{masa}, \ac{mmse} and \ac{dlased} performed better with \ac{pnn} and \ac{tead} got worse.

Considering results from all three models\textsuperscript{\ref{github}} for the 4D benchmark functions, we found that \ac{masa} provided the highest $R^2$ ($0.849$) and $R^2\sub{Area}$ ($0.724$) score across all models on average. Furthermore, \ac{guess} achieved the highest rank regarding $R^2$, second best $R^2$ score ($0.842$), and third highest rank regarding $R^2\sub{Area}$. 
\ac{mepe} showed the weakest performance with average $R^2$ of $0.83$ and $R^2\sub{Area}$ of $0.714$ across all models.
\subsection{\change{Computational Effort}}\label{secA_comp}
\begin{figure*}[!tb]%
	\centering
	\includegraphics[width=1.0\textwidth]{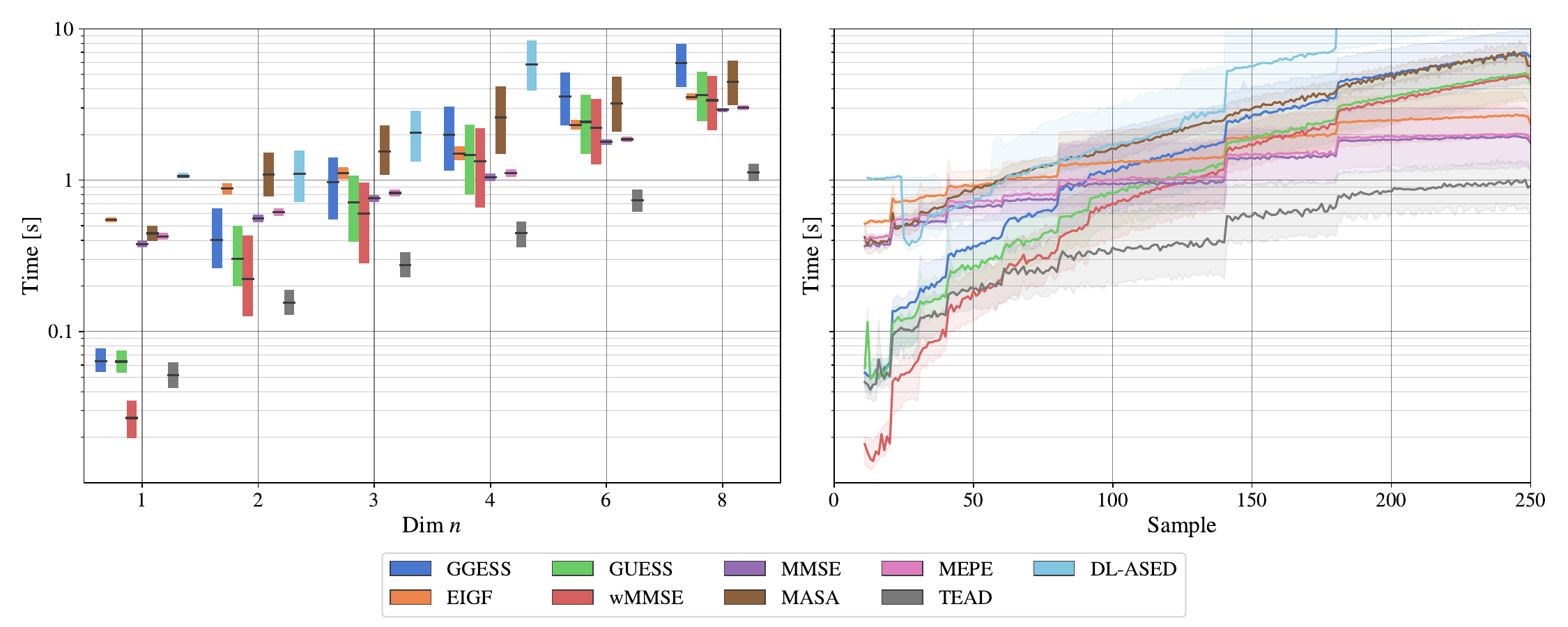}
	\caption{\change{Comparison of wall-clock time [s] for proposing one new sample point (Step 4 in Section \ref{sec3_1}) for different adaptive sampling strategies with \ac{gp} surrogate model for the 1 to 8-dimensional benchmark study (Section \ref{sec4}). Left: \ac{iqd} of the time across dimensions; boxes indicate the 25\% and 75\% quantile and the horizontal line the median. Right: \ac{iqd} of the time across proposed samples; filled area indicate the 25\% and 75\% quantile and the solid line the median. \ac{dlased} is not fully visible on the left plot for 6D (median: $39$ seconds, 25\%-quantile: $36$ seconds, 75\%-quantile: $41$ seconds) and on the right (median: $18$ seconds at 250 samples). For 8D no results for \ac{dlased} are shown, due to sampling of the edges.}}\label{figtiming}
\end{figure*}

\change{Computational effort besides evaluating the black-box function is mainly driven by the expense of (re-)training the surrogate model (Step 3 in Section \ref{sec3_1}), and evaluation of $\phi$ for optimization or assessing the candidate points (Step 4 in Section \ref{sec3_1}). Model training is almost identical between adaptive strategies and independent of the acquisition function, except for \ac{masa} which has to update the whole ensemble. Therefore, we show in Figure \ref{figtiming} a comparison of the wall-clock time for proposing one new sample point (Step 4 in Section \ref{sec3_1}) with the different sampling strategies for the 1 to 8-dimensional benchmark study (Section \ref{sec4}). Benchmarks are run parallelized over all 10 repetitions on a desktop computer with Intel i7-6900k and 48 GB DDR-4 computer memory. The plot is limited to 10 seconds to improve visibility. Therefore, \ac{dlased} is not fully visible on the left for 6D (median: $39$ seconds, 25\%-quantile: $36$ seconds, 75\%-quantile: $41$ seconds) and on the right (median: $18$ seconds at 250 samples). Moreover, for \ac{dlased} we used only samples proposed by the acquisition function, \ie we removed samples placed by edge sampling (see Section \ref{sec4_1}). Therefore, no results for 8D are shown for \ac{dlased}, due to exhausting sampling of the edges, \ie sampling edges exceeded maximum number of samples.}

\change{Proposing new samples becomes slower with increasing sample size, partially since \ac{gp} scales $\mathcal{O}(m^3)$ with sample size. It can be seen that with increasing dimensionality of the benchmark function, the proposal time also increases due to the increase in sample size. However, it should be noted the maximum number of samples was limited for 6 and 8-dimensional benchmarks to $250$. As can be seen, \ac{tead} needs comparably low compute for proposing new samples. We found the usage of model uncertainty in the acquisition function to be one of the main driver of computational complexity.}

\change{We can further improve the wall time by decreasing the model training frequency. A commonly used strategy for \ac{gp} would be to train the surrogate model only every $i$-th iteration. In the remaining iterations, the \ac{gp} can be conditioned on new observations, \ie we would build a model including the new observations using the model parameters $\boldsymbol{\theta}\sub{GP}$ from the previous iteration. Moreover, since computational effort was not the focus of this study, more efficient implementation of the strategies may improve computation further. For sampling in high dimensions or larger sample sizes, one could also consider approaches mentioned in \ref{secA_challenges}.}

\section{Implementation Details}\label{secB}
Our code was implemented in Python \citep{vanrossum1995} with several different packages and is available on GitHub\textsuperscript{\ref{github}} for further research. The \ac{gp} used in this work is based on the implementation in the scikit-optimize library \citep{tim2021} which relies on the library scikit-learn \citep{pedregosa2011}. For training of the \ac{gp} the default optimizer based on scipy's \citep{virtanen2020} implementation of the L-BFGS-B algorithm \citep{byrd1995} was used with 10 random restarts to maximize the log-marginal likelihood. A small value $(10^{-10})$ is added to the diagonal of the kernel matrix during fitting to provide numerical stability. A \ac{pnn} implementation based on \citep{chua2018} was used with two dif\-ferent net topologies and $n_{en}=5$. A network with one layer, 64 neurons and Swish activation function were used for \Back{4}, \Brose{4} and \Bstyb{4}. Furthermore, a network with one layer, 48 neurons and Tanh activation function were used for \Bmicha{4}. Adam \citep{kingma2015} was used throughout with a learning rate of $0.03$ to minimize the loss function (Eq. \eqref{eq_pnn_loss}). \ac{dgcn} is a custom implementation based on \citep{cremanns2021} which uses TensorFlow \citep{martinabadi2015} as backend and five different kernel functions as described in \citep{cremanns2021}. \change{For \ac{svgp}, we used the implementation from \citep{gpy2014}.}

Two evolutionary strategies were used for optimizing the continuous acquisition function in Eq. \eqref{eq17}, as implemented in pygmo \citep{biscani2020} (for \ac{gp}) and with the differential evolution algorithm in scipy (for \ac{dgcn} and \ac{pnn}).  The baseline \ac{lhs} used in the benchmarks is a custom implementation based on \cite{bogoclu2021}, where samples are drawn from $\VSpace{X}$ without correlation and by maximizing pairwise distance.  Additionally, the implementation in the scikit-optimize library \citep{tim2021} was used for all other use cases. $R_{Area}^2$ in Eq. \eqref{eq37} was calculated with the composite Simpson' rule from the library scipy \citep{virtanen2020}. The parameter ‘even=last’ has to be passed at the function call together with a vector with $s$ evenly spaced values on $[0, 1]$ to yield the same results as explained in Eq. \eqref{eq37}, including start and end points. 

\section{Benchmark Functions}\label{secC}
Table \ref{tabC1} gives an overview of the 15 different benchmark functions  used throughout this study. The last five functions can be adapted to different dimensions.
\begin{table*}[!th]
	\begin{center}
		\begin{minipage}{\textwidth}
			\caption{Benchmark functions}\label{tabC1}
			\footnotesize
			\centering
			\begin{tabular}[t]{@{}M{4.2cm} M{3cm} M{7.8cm}@{}}
				\toprule
				\textbf{Name} & \textbf{Input}  & \textbf{Function}\\
				\midrule
				\Bhsingle (modified) & $x \in [-1.5, 5]$  &
				$\begin{aligned}[t]
					f\sub{HS}(\textbf{x})=\frac{0.05}{{\left(x-4.75\right)}^{2}+0.004} -\frac{0.09}{{\left(x-4.45\right)}^{2}+0.05} -6 + 3x
				\end{aligned}$
				\\[3.5ex]
				
				\Bhtwo (modified) & $x \in [-0.5, 5]$  & 
				$\displaystyle f\sub{HT} (\textbf{x})=5x+\frac{0.05}{{\left(x-4.5\right)}^{2}+0.002}-
				\frac{0.5}{{\left(x-3.5\right)}^{2}+3.5}-6 $ \\[3.5ex]
				
				\Bgram \citep{gramacy2010}     & $x \in [-1.5, 1]$  & $\displaystyle f\sub{GL} (\textbf{x})= \frac{60\mathrm{sin}\left(6\pi x\right)}{2\mathrm{cos}\left(x\right)}+{\left(x-1\right)}^{4}$ \\[2.5ex]
				
				\Bbeker \cite{ajdari2014}   & 
				$
				\begin{aligned}
					&x_d \in [-10, 10], \\ &d=1,2
				\end{aligned}$
				& 
				$\displaystyle
				f\sub{BL} (\textbf{x})=\left(\left\vert x_1 \right\vert -5 \right)^{2} +\left(\left\vert{x}_{2}\right\vert-5\right)^{2}
				$\\[2.5ex]
				
				\Begg \citep{mishra2006}   & 
				$
				\begin{aligned}[t]
					&x_d \in [-512, 512], \\&d=1,2
				\end{aligned}$  
				& 
				$\begin{aligned}[t]
					f\sub{EGG} (\textbf{x})=&-(x_2+47)\sin\left(\sqrt{\left\vert x_2+0.5x_1+47\right\vert}\right) \\&-x_1 \sin \left(\sqrt{\left\vert x_1-(x_2+47)\right\vert}\right)
				\end{aligned}$
				\\[4.5ex]
				
				\Bhimm \citep{himmelblau1972}   & 
				$
				\begin{aligned}
					&x_d \in [-6, 6], \\&d=1,2
				\end{aligned}
				$  
				
				& $f\sub{HIM} (\textbf{x})=(x_1^2+x_2-11)^2+(x_1+x_2^2-7)^2$ \\[2.5ex]
				
				\Bbran \citep{branin1972}   &
				$
				\begin{aligned}
					&x_d \in [-5, 10], \\&d=1,2
				\end{aligned}
				$
				& 
				$\begin{aligned}[t]
					f\sub{BRN}(\textbf{x})=\left(x_2- \frac{5.1}{4\pi^2}x_1^2+\frac{5}{\pi}x_1-6\right)^2
					+10\left(1-\frac{1}{8\pi}\right)\cos(x_1)+10
				\end{aligned}$
				\\[3.5ex]
				
				\Bdrop \cite{contreras2014}   & 
				$
				\begin{aligned}
					&x_d \in [-0.6, 0.9], \\&d=1,2
				\end{aligned}     
				$                           
				& 
				$\begin{aligned}[t]
					f\sub{DRP}(\textbf{x})=-\left(1+ \cos\left(12\sqrt{x_1^2+x_2^2}\right)\right)
					\left(0.5(x_1^2+x_2^2)+2\right)^{-1}
				\end{aligned}$
				\\[2.5ex]
				
				\Bishi \citep{ishigami1991}   & 
				$
				\begin{aligned}
					&x_d \in [-\pi, \pi], \\&d=1,2,3
				\end{aligned}
				$
				& $f\sub{ISH}(\textbf{x})=\sin(x_1)+7 \sin^2(x_2)+0.1x_3^4\sin(x_1)$\\[2.5ex]
				
				\Bhart\citep{hartman1973}   & 
				$
				\begin{aligned}
					&x_d \in [0, 1], \\&d=1,...,6
				\end{aligned}
				$
				& 
				$\begin{aligned}[t]
					&f\sub{HRT}(\textbf{x})=-\sum^4_{i=1} \boldsymbol{\alpha}_i \exp\left( -\sum^6_{j=1} \textbf{A}_{ij}(x_j-\textbf{P}_{ij})^2\right),\\
					&\text{where}\quad 
					\boldsymbol{\alpha}=(1, 1.2, 3, 3.2)^T, \\
					& \textbf{A}=\begin{pmatrix}
						10 & 3 & 17 & 3.5 & 1.7 & 8 \\
						0.05 & 10 & 17 & 0.1 & 8 & 14 \\
						3 & 3.5 & 1.7 & 10 & 17 & 7 \\
						17 & 8 & 0.05 & 10 & 0.01 & 14
					\end{pmatrix},\\
					&\textbf{P}=10^{-4}\begin{pmatrix}
						1312 & 1696 & 5569 & 124 & 8283 & 5886 \\
						2329 & 4135 & 8307 & 3736 & 1004 & 9991 \\
						2348 & 1451 & 3522 & 2883 & 3047 & 6650 \\
						4047 & 8828 & 8732 & 5743 & 1091 & 381 \\
					\end{pmatrix}
				\end{aligned}$
				\\[31.5ex]
				
				\Brose[]{}\citep{rosenbrock1960}   & 
				$
				\begin{aligned}
					&x_d \in [-5, 5], \\&d=1,...,\ndim
				\end{aligned}
				$
				& 
				$\begin{aligned}[t]
					f\sub{ROS}(\textbf{x})=-\sum_{i=1}^{\ndim-1} [ 100(x_{i+1}-x_i^2)^2 +(x_i-1)^2 ]
				\end{aligned}$
				\\[3.5ex]
				
				\Back[]{}\citep{ackley1987}   & 
				$
				\begin{aligned}
					&x_d \in [-5, 5], \\&d=1,...,\ndim
				\end{aligned}
				$
				& 
				$\begin{aligned}[t]
					f\sub{ACL}(\textbf{x})=& -20\exp\left(-0.2\sqrt{\ndim^{-1}\sum^{\ndim}_{i=1}x_i^2}\right)
					-\exp\left(\ndim^{-1}\sum^{\ndim}_{i=1}\cos(2\pi x_i)\right)\\&+20+\exp(1)
				\end{aligned}$
				\\[10ex]
				\Bmicha[]{}\citep{michalewicz1992}   & 
				$
				\begin{aligned}
					&x_d \in [0, \pi], \\&d=1,...,\ndim
				\end{aligned}
				$
				& $\begin{aligned}f\sub{MIC}(\textbf{x})=-\sum^{\ndim}_{i=1}\left[ \sin(x_i) \sin^{10}\left(\frac{i x_i^2}{\pi} \right) \right]\end{aligned}$\\[3ex]
				
				\Bschw[]{}\citep{schwefel1981}   &
				$
				\begin{aligned}
					&x_d \in [-5, 5], \\&d=1,...,\ndim
				\end{aligned}
				$                                
				& $\begin{aligned}f\sub{SCH}(\textbf{x})=418.9829 \ndim-\sum^{\ndim}_{i=1}x_i \sin\left( \sqrt{x_i} \right)\end{aligned}$\\[3.5ex]
				
				\Bstyb[]{}\citep{styblinski1990}  &
				$
				\begin{aligned}
					&x_d \in [-5, 5], \\&d=1,...,\ndim
				\end{aligned}
				$                                
				&
				$\begin{aligned}f\sub{STY}(\textbf{x})= 0.5 \sum_{i=1}^{\ndim} \left( x_i^4 - 16x_i^2 + 5x_i \right)\end{aligned}$\\
				
				\bottomrule
			\end{tabular}
		\end{minipage}
	\end{center}
\end{table*}

\end{document}